\documentclass{article}

\usepackage{microtype}
\usepackage{graphicx}
\usepackage{booktabs} % for professional tables
\usepackage{amsfonts}
\usepackage{hyperref}
\usepackage{amsmath}
\usepackage{cleveref}
\usepackage{xspace}
\usepackage{subcaption}

\newcommand{\algname}{AccelOpt\xspace}
\newcommand{\benchname}{NKIBench\xspace}
\usepackage{listings}
\lstdefinestyle{prompt}{
  basicstyle=\ttfamily\tiny,breaklines=true,columns=fullflexible,
  frame=single,framesep=6pt
}
\usepackage{diagbox}
\usepackage{amsmath} % for \DeclareMathOperator*

\usepackage{pifont}

\usepackage[accepted]{mlsys2025}

\mlsystitlerunning{AccelOpt: A Self-Improving LLM Agentic System for AI Accelerator Kernel Optimization}

\begin{document}

\twocolumn[
\mlsystitle{AccelOpt: A Self-Improving LLM Agentic System for AI Accelerator Kernel Optimization}

% unless you've provided the [accepted] option to the mlsys2025


\mlsyssetsymbol{past}{$\dagger$}

\begin{mlsysauthorlist}
\mlsysauthor{Genghan Zhang}{stf,past}
\mlsysauthor{Shaowei Zhu}{aws}
\mlsysauthor{Anjiang Wei}{stf,past}
\mlsysauthor{Zhenyu Song}{aws,past}
\mlsysauthor{Allen Nie}{aws,past}
\mlsysauthor{Zhen Jia}{aws}
\mlsysauthor{Nandita Vijaykumar}{aws,uot}
\mlsysauthor{Yida Wang}{aws}
\mlsysauthor{Kunle Olukotun}{stf}
\end{mlsysauthorlist}

\mlsysaffiliation{stf}{Department of Computer Science, Stanford University, USA}
\mlsysaffiliation{aws}{Amazon Web Services, USA}
\mlsysaffiliation{uot}{Department of Computer Science, University of Toronto, Canada}

\mlsyscorrespondingauthor{Genghan Zhang}{zgh23@stanford.edu}

\mlsyskeywords{Machine Learning, MLSys}

\vskip 0.3in

\begin{abstract}
We present AccelOpt, a self-improving large language model (LLM) agentic system that autonomously optimizes kernels for emerging AI acclerators, eliminating the need for expert-provided hardware-specific optimization knowledge.
AccelOpt explores the kernel optimization space through iterative generation, informed by an optimization memory that curates experiences and insights from previously encountered slow-fast kernel pairs.
We build NKIBench, a new benchmark suite of AWS Trainium accelerator kernels with varying complexity extracted from real-world LLM workloads to evaluate the effectiveness of AccelOpt.
Our evaluation confirms that AccelOpt's capability improves over iterations, boosting the average percentage of peak throughput from $49\%$ to $61\%$ on Trainium 1 and from $45\%$ to $59\%$ on Trainium 2 for NKIBench kernels. Moreover, AccelOpt is highly cost-effective: using open-source models, it matches the kernel improvements of Claude Sonnet 4 while being $26\times$ cheaper. The code is open-sourced at \url{https://github.com/zhang677/AccelOpt}.
\end{abstract}
]



\printAffiliationsAndNotice{
    \textsuperscript{$\dagger$}Part of the work done while interning or working at AWS.
} % leave blank if no need to mention equal contribution
% \printAffiliationsAndNotice{\mlsysEqualContribution} % otherwise use the standard text.

\section{Introduction}
\label{sec:intro}
\begin{figure*}[thb]
\begin{center}
\centerline{\includegraphics[width=1.8\columnwidth]{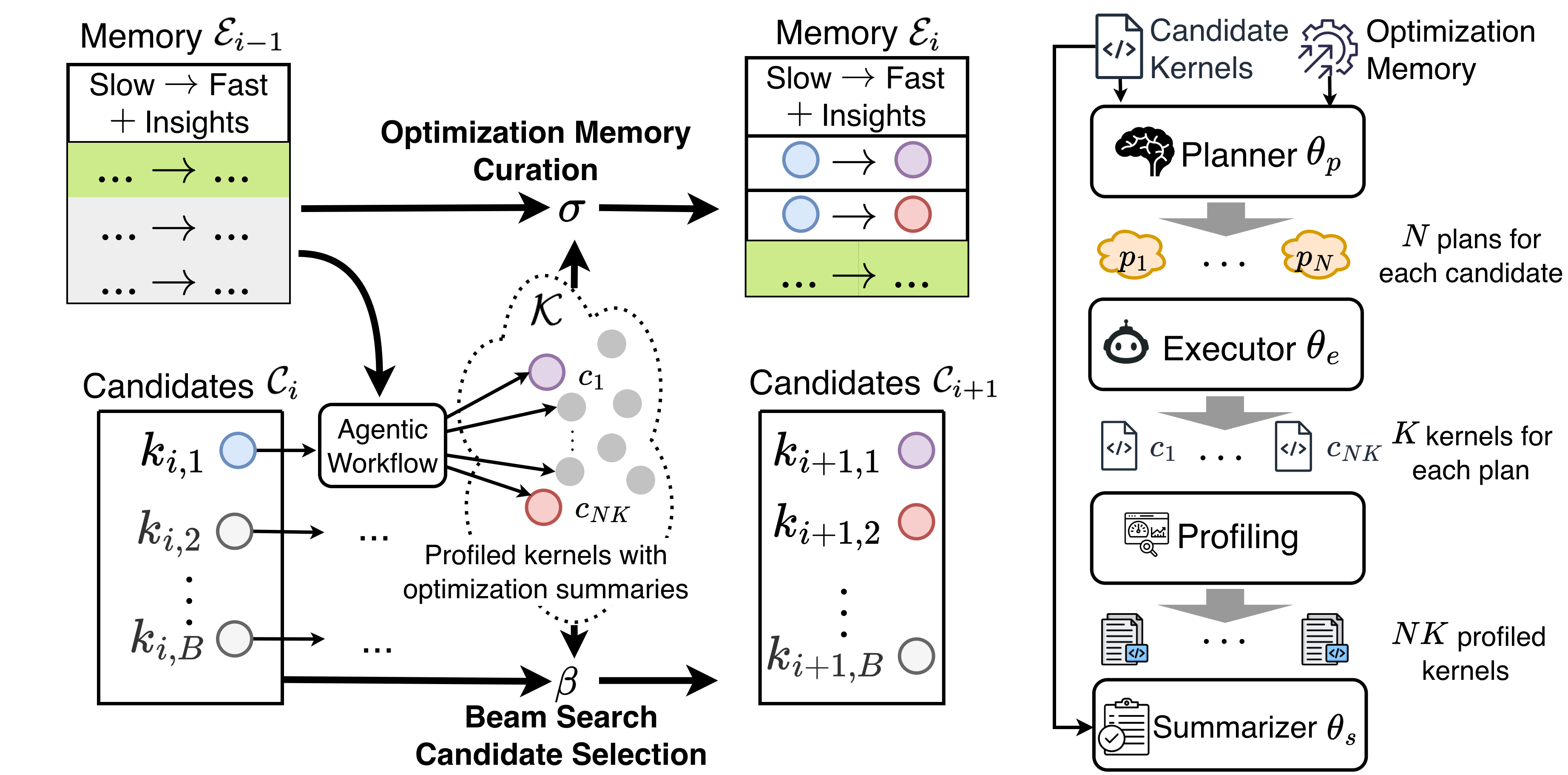}}
\vspace*{-2ex}
\caption{At each iteration of \algname, the agentic workflow shown on the right optimizes the candidate kernels with the latest optimization memory, and generates new candidate kernels, updating optimization memory with newly collected experiences. ~\Cref{sec:algorithm} explains the overall workflow and each component in detail.}
\label{fig:main-method}
\end{center}
\end{figure*}

The unprecedented demand for compute power in the age of large models has prompted the rise of AI accelerators~\cite{abts2022groq,lie2022cerebras,jouppi2023tpu,prabhakar2024sambanova,aws-trainium}. However, their performance critically depends on the efficiency of kernels, the low-level implementations that determine how machine learning operators are mapped onto hardware resources. Suboptimal kernels can severely limit system performance and, when scaled to large deployments, result in substantial waste of compute and financial resources~\cite{spector2024thunderkittens,ye2025flashinfer,zhao2025insights}.

Kernel optimization, however, is notoriously difficult and demanding, even for well-understood architectures like GPUs. 
For instance, after NVIDIA released H100 in 2022, it took about a year for attention kernels to reach roughly 37\% of theoretical peak performance~\cite{dao2023flashattention} and another year to approach 85\% \cite{shah2024flashattention}.
% given that attention dominates today’s LLM workloads—the primary workloads on cloud GPUs. 
Achieving high efficiency requires navigating a complex interplay between workload characteristics, memory hierarchies, parallelism, and architecture-specific constraints. As a result, empirical tuning and extensive exploration of the optimization space are necessary for producing efficient kernels~\cite{jia2019taso,wu2025mirage}.
The challenge is even greater for emerging AI accelerators, whose architectures diverge significantly from GPUs, leaving kernel developers with limited performance intuition and few established optimization heuristics \cite{hsu2025stardust,fang2025dato}.

In this work, we focus on improving kernels for Amazon Trainium~\cite{aws-trainium}, a widely deployed and representative AI accelerator that exemplifies these challenges. 
Trainium is programmed with Neuron Kernel Interface (NKI), a Python-embedded kernel language, where both the hardware and the programming model remain relatively new~\cite{nki_release}. Consequently, developers lack the extensive optimization recipes and performance heuristics available for mature platforms like GPUs~\cite{Thakkar_CUTLASS_2023}. 
This makes kernel optimization a critical, real-world challenge faced by engineers today for every new accelerator entering production~\cite{openai_chip,qual_chip,meta_chip,ms_chip}. 

Moreover, cost efficiency becomes increasingly important at scale. As AI accelerators proliferate and workloads diversify, practitioners face hundreds or thousands of kernels to optimize across different configurations, hardware versions, and workloads~\cite{kim2023full}. In this regime, achieving expert-level results in a cost-efficient way enables broader adoption of the accelerator and faster iteration cycles.

Therefore, we explore the potential of using LLMs to generate optimized kernels for the Trainium accelerator. 
LLMs have shown potential to automatically generate correct kernels with competitive performance in the context of GPUs, TPUs, and NPUs~\cite{ouyang2025kernelbench,wei2024improving,li2025cuda,lange2025towards,agrawal2025gepa,hong2025autocomp,novikov2025alphaevolve,woo2025tritonrl}.
In the context of Trainium, given the limited availability of Trainium-specific optimization knowledge and kernel tuning recipes, an important goal of this work is to investigate whether an LLM-based system can autonomously navigate the optimization space to produce high-performance kernels without relying on human-engineered heuristics or preexisting optimization examples.

This task poses two challenges.
First, similar to other AI accelerators, Trainium kernel optimization necessitates exploring a vast design space encompassing memory layouts, parallelization schemes, and scheduling strategies. 
However, since LLM queries incur substantial computational costs, this exploration must be conducted strategically to balance search space coverage with cost efficiency. 
Second, we aim to enable the LLM-based system to autonomously accumulate optimization insights during such explorations, allowing the system to progressively improve its capabilities over time without requiring manual intervention.   

To address these challenges, we propose \algname, a self-improving LLM agentic system, which utilizes beam search with optimization memory on top of an agentic workflow. 
\algname uses \emph{beam search} to explore the Trainium kernel optimization space through iterative generation of new kernels based on old ones, while retaining top-performing candidate kernels for consideration.
When generating new kernels in each iteration, \algname employs a three-component agentic workflow—planner, executor, and summarizer—that mimics how human experts approach the problem. 
Profiles of the generated kernels are obtained through a distributed profiling service and subsequently used to curate an \emph{optimization memory}. 
This memory stores a selection of past exploration experiences, including key code changes that resulted in slow-to-fast kernel transformations, along with LLM-generated summaries of general optimization insights. 
The optimization memory then serves to inspire new optimization strategies in future iterations.
To comprehensively evaluate \algname, we construct \benchname, a benchmark suite containing challenging kernels from real-world LLM workloads. 
A distinguishing feature of this benchmark is that we calculate the theoretical peak performance achievable by the hardware for each task. 
This enables us to assess the system's position within the entire kernel optimization landscape, providing deeper insights beyond simply measuring relative speedup compared to the initial kernel.

The evaluation of \algname shows promising results and interesting insights.
We observe that \algname is able to navigate the Trainium kernel optimization space by discovering both local optimizations and non-trivial global optimizations~(\Cref{sec:opt-case}). 
We also explore the cost-benefit trade-off in \algname, where a large amount of kernels are sampled using LLMs, which can incur non-trivial cost, by evaluating the system under different optimization memory configurations and base LLMs~(\Cref{sec:cost-analysis}).
Using open-source LLMs, \algname improves the average percentage of peak throughput from $49\%$ to $61\%$ on Trainium 1, which is on par with Claude Sonnet 4 (thinking mode) but 26$\times$ cheaper, and from $45\%$ to $59\%$ on Trainium 2 at a similar cost across all tasks in \benchname. 

This work makes the following contributions:
\begin{itemize}
    \item We propose \algname, the first self-improving LLM agentic system for kernel optimization on emerging AI accelerators that combines search with memory accumulation. \algname is the first system that does not require expert-provided, hardware-specific optimization knowledge or predefined optimization recipes, among the open-source systems we are aware of.
    \item We construct \benchname, the first benchmark suite for NKI kernel optimization on Amazon Trainium, with all kernels derived from real-world LLM workloads. \benchname measures kernel performance against theoretical peak hardware performance on Trainium, rather than relying solely on relative speedup metrics, which can be ambiguous due to different baseline choices.
    \item We demonstrate that \algname discovers substantial optimizations on real-world kernels in \benchname. The 14 kernels in the current version of \benchname establish a starting point for future NKI kernel optimization research. Furthermore, we show that \algname can leverage open-source LLMs (gpt-oss-120b and Qwen3-Coder-480B-A35B-Instruct-FP8) to attain comparable performance improvements at significantly lower cost than those achieved using Claude Sonnet 4, one of the leading proprietary models for code generation.
    \item We verify that beam search is a more effective inference-time scaling technique than repeated sampling. Further, we find that including optimization memories with beam search affects cost efficiency in obtaining good-performing NKI kernels in various ways. However, it does not significantly improve the performance of the best kernels discovered if enough kernels are sampled, compared to beam search alone.
\end{itemize}

\section{AccelOpt}
\label{sec:algorithm}
We will first introduce the overall architecture of \algname in~\Cref{sec:general-algorithm}, before diving into the details on the design of beam search~(\Cref{sec:beam-search}) and optimization memory~(\Cref{sec:memory-design}) components.
\subsection{Algorithm Overview}
\label{sec:general-algorithm}
% At a high level, \algname contains LLM agents operating in a beam search while maintaining optimization memory of past experiences. Each iteration progressively expands the search frontier by generating new kernel implementations while consolidating what agents have learned through profiling and summarizing into the optimization memory to inform future iterations.

The key insight of \algname is to let the agents explore and learn from their own optimization experience. 
Two mechanisms make this possible: \textbf{beam search}, which iteratively updates the frontier of candidate kernels and surfaces the best ones for the next round of exploration; \textbf{optimization memory}, which contains distilled optimization insights and key code changes from discovered slow-fast kernel pairs and transfers them to future iterations.

\begin{algorithm}[tbh]
\caption{Candidate kernel frontier expansion and memory update in one iteration of \algname. }
\label{alg:nkiopt}
\begin{algorithmic}[1]
\INPUT $\mathcal{E}_{i-1}$: experience at iteration $i-1$; $\mathcal{C}_i$: candidate kernels at iteration $i$, $|\mathcal{C}_i|=B$
\REQUIRE $\theta_p$: planner, $\theta_e$: executor, $\theta_s$: summarizer, $r$: profiler function, $\sigma$: optimization memory curation, $\beta$: candidate selection function
\STATE $\mathcal{K} \leftarrow \emptyset$
\FOR{$c\in \mathcal{C}_i$}
\STATE $\mathcal{P} = \{p\mid p\sim \theta_p(p\mid c, \mathcal{E}_{i-1})\}$  \hfill $\triangleright|\mathcal{P}|=N$ 
\FOR{$p \in \mathcal{P}$}
\STATE $\mathcal{A}_p = \{(a, p, r(a)) \mid a\sim \theta_e(e\mid p, c)\}$ 
\STATE $\mathcal{K} = \mathcal{K} \cup \mathcal{A}_p$ \hfill $\triangleright|\mathcal{A}_p|=K$ 
\ENDFOR
\ENDFOR
\STATE $\mathcal{E}_i = \sigma(\mathcal{K}, \mathcal{E}_{i-1}; \theta_s)$ \hfill $\triangleright$ See \Cref{alg:exp-gen}
\STATE $\mathcal{C}_{i+1} = \beta(\mathcal{K}\cup \mathcal{C}_i, B)$
\OUTPUT $\mathcal{K}$, $\mathcal{E}_i$, $\mathcal{C}_{i+1}$
\end{algorithmic}
\end{algorithm}

\algname agentic workflow responsible for generating new kernels from old consists of three interacting agents, as shown on the right of \Cref{fig:main-method}. 
\Cref{fig:plan-example} shows a snapshot of an \algname execution trace. In this example, the \textit{planner} uses profiling results to identify memory operations as a performance bottleneck and proposes eliminating redundant computation accordingly. Guided by the plan, the \textit{executor} performs kernel optimizations involving multi-level loop transformations and tensor layout changes. The \textit{summarizer} then distills a generalizable optimization strategy, namely ``reusing precomputed results'', and optimization segments of the slow-fast pairs. ~\Cref{fig:short-prompts} shows the prompt template for each agent and prompt details are in Appendix~\Cref{sec:appendix-prompts}. 

\begin{figure}[tbh]
    \centering
    \includegraphics[width=\linewidth]{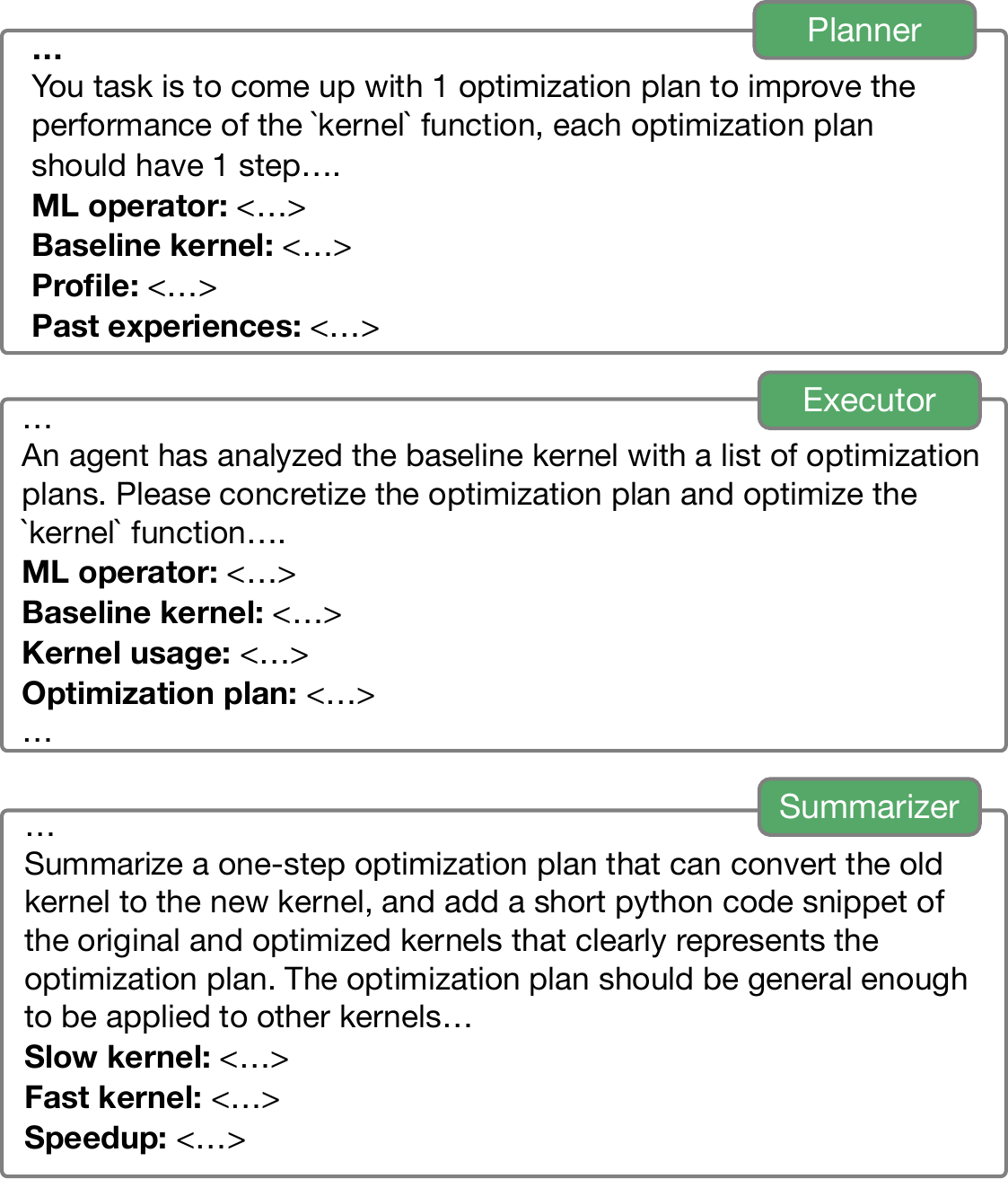}
    \caption{Prompt template for each agentic in the agentic workflow.}
    \label{fig:short-prompts}
\end{figure}

\subsection{Beam Search}
\label{sec:beam-search}
\citet{hong2025autocomp} has shown that beam search can effectively lead to performance improvements on LLM-generated kernels for some hardware accelerators.
We also adopt this mechanism in this work, and we confirm that it is a more effective method compared to the simple parallel (repeated) sampling for open-source LLMs through our experiments.

As shown in~\Cref{alg:nkiopt} and \Cref{fig:main-method}, at each iteration $i$, the planner agent generates $N$ plans for each kernel in a set of $B$ candidate kernels augmented with experiences from iteration $i-1$. After that, the executor agent implements every plan with $K$ attempts, generating $B\times N\times K$ kernels in total. By sampling multiple plans for the same candidate, the planner explores diverse optimization strategies, and multiple executor attempts increase the robustness of plan implementation against syntactic and semantic errors. From these generated kernels, high-quality optimizations are selected for the summarizer agent to generate experience items, which are used in the curation of the optimization memory. Finally, $B$ kernels are selected to be explored in the next iteration from those $(B + B\times N\times K)$ kernels.

Central to the beam search algorithm, the candidate selection function $\beta$ is responsible for selecting the $B$ candidates to continue exploring in the next iteration.
$\beta$ first identifies the fastest correct kernel within each plan group $\mathcal{A}_p$, ensuring that every explored direction contributes its best result. From this representative pool, it then selects the top-$B$ kernels by measured latency. If fewer than $B$ valid kernels exist, remaining slots are filled by the previous iteration’s candidates, allowing the system to dynamically allocate more sampling budget to difficult cases where no improvement was achieved. 

\subsection{Optimization Memory Curation}
\label{sec:memory-design}
\begin{figure*}[thb]
\begin{center}
\centerline{\includegraphics[width=2\columnwidth]{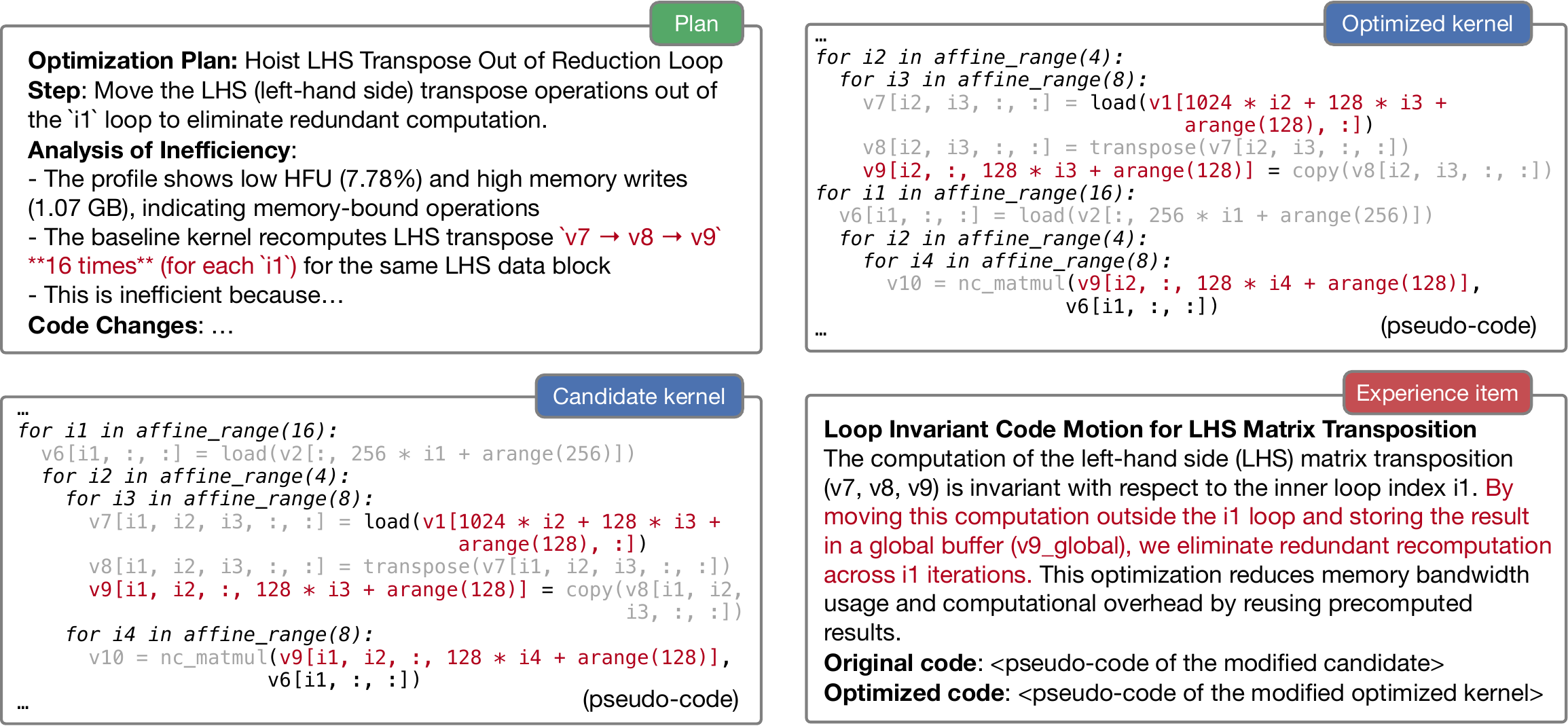}}
\vspace*{-1ex}
\caption{A snapshot of \algname's execution trace. In the experience item, the pseudocode of the slow-fast pairs looks like the above candidate and optimized kernels where \texttt{affine\_range} is a NKI construct for parallel loops without carried dependency. The experience item will be stored in the optimization memory, and the optimized kernel will become a candidate for the next iteration.}
\label{fig:plan-example}
\end{center}
\end{figure*}
Although beam search can record exploration history through its evolving candidates, it cannot capture optimization experiences. Therefore, we design optimization memory curation, which collects optimization insights during exploration. This optimization memory expands the knowledge of the accelerator’s optimization space for both LLM agents and humans, and enhances exploration efficiency.

\begin{algorithm}[tbh]
\caption{Optimization memory curation procedure $\sigma$.}
\label{alg:exp-gen}
\begin{algorithmic}[1]
\INPUT $\mathcal{K}$, $\mathcal{E}_{i-1}$
\REQUIRE $\theta_s$, TopK, ExpN, $t_{pos}$, $t_{neg}$
\STATE $\mathcal{R}_{pos}\leftarrow \emptyset$, $\mathcal{R}_{neg}\leftarrow \emptyset$
\STATE // Group by candidates and plans for each kernel
\STATE $\mathcal{S} = \mathcal{K}$.groupby(c, p)
\FOR{$s_{c,p} \in S$}
\IF{$s_{c,p}$.max\_speedup $>t_{pos}$}
\STATE $\mathcal{R}_{pos}$.add((c, $s_{c,p}$.fastest\_kernel))
\ELSIF{$s_{c,p}$.max\_speedup $<1 / t_{neg}$}
\STATE $\mathcal{R}_{neg}$.add(($s_{c,p}$.slowest\_kernel, c))
\ENDIF
\ENDFOR
\STATE $\mathcal{E}_{pos} = \bigl[ \theta_s(r) \mid r \in \mathcal{R}_{pos}.sort()[:\text{TopK}//2]\bigr]$
\STATE $\mathcal{E}_{neg} = \bigl[\theta_s(r) \mid r \in \mathcal{R}_{neg}.sort()[:\text{TopK}-|\mathcal{E}_{pos}|]\bigr]$
\STATE $\mathcal{E}_{i+1} = \bigl[\mathcal{E}_{pos}, \mathcal{E}_{neg}, \mathcal{E}_i[:\text{ExpN}-|\mathcal{E}_{pos}| - |\mathcal{E}_{neg}|]\bigr]$
\OUTPUT $\mathcal{E}_{i+1}$
\end{algorithmic}
\end{algorithm}

As shown in Algorithm~\ref{alg:exp-gen}, the optimization memory is maintained as a queue of optimization items with a capacity cap (ExpN).
Each new iteration can append up to TopK experience items to the tail, while the oldest entries in the memory will be discarded once ExpN is reached.
Intuitively, increasing ExpN can
retain more historical experiences that can potentially be beneficial. 
The TopK parameter controls how eager the memory system can be when updating the memory using the current iteration observations. Higher TopK and ExpN can both lead to higher inference costs.
We provide a cost-benefit analysis of these parameters in \Cref{sec:evaluation}.

Each experience item in the optimization memory consists of a slow-fast kernel pair and the corresponding generalizable optimization strategy curated by the summarizer agent. To prevent irrelevant code from distracting the planner, the summarizer extracts the optimized segment of each pair as pseudocode. Slow-fast pairs come from two sources: (1) the baseline kernel and a generated faster kernel (positive rewrites), and (2) a generated slower kernel and the baseline kernel (negative rewrites). Both positive and negative rewrites represent performance-improvement cases. One highlights successful optimization, and the other captures failed attempts. Therefore, we include both to provide balanced signals for the self-improving system. To ensure quality, $\sigma$ applies speedup thresholds $t_{\text{pos}}$ and $t_{\text{neg}}$ to them, respectively. To encourage diversity, $\sigma$ groups kernels by their originating candidates and plans, selecting performance outliers within each subgroup.

AccelOpt's memory component is part of the test-time learning paradigm~\cite{sun2024learning} that applies on a per-problem basis for iterative kernel optimization, where optimization insights transfer from previous iterations to future ones. Studying how to transfer the memory accumulated from optimizing some kernels to others is worth future exploration.

\section{Benchmarks and Evaluation Infrastructure}
\begin{figure}[thb]
    \centering
    \includegraphics[width=\linewidth]{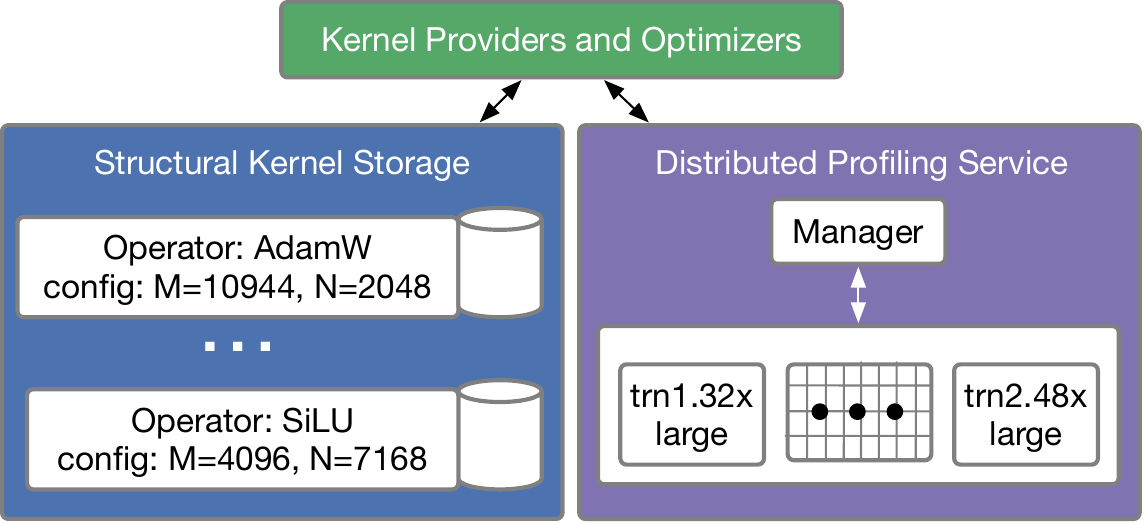}
    \vspace*{-3ex}
    \caption{\benchname architecture. Kernels are grouped by the configuration of ML operators. Each kernel instance stores both the kernel code and profiling information. The meshes represent cores of one Trainium chip; trn1.32xlarge and trn2.48xlarge are Amazon EC2 instances for Trainium 1 and 2, respectively.}
    \label{fig:benchmark-flow}
\end{figure}

At the time of developing \algname, no existing benchmark suite contained NKI kernels with sufficient baseline performance to serve as meaningful starting points for optimization. 
Moreover, existing accelerator kernel benchmarks typically lack information about how well a kernel is optimized relative to the hardware's theoretical peak performance. To address these gaps, we construct \benchname, which provides challenging kernel optimization tasks from real-world LLM workloads managed by a structural storage together with a distributed kernel profiling service that enables efficient execution and evaluation of \algname at scale, as shown in~\Cref{fig:benchmark-flow}.

\subsection{\benchname Task Construction}
\benchname initial kernels were primarily generated by the official Neuron compiler, which provides a reasonable baseline. We collect 14 representative NKI kernels from popular LLM workloads with reasonable initial performance (\Cref{fig:ratio-bars}). For four problems where the compiler did not provide implementations, we manually created kernels optimized using standard techniques (tiling, fusion) or adapted kernels from official NKI examples. One of the kernels is from a non-transformer LLM, and the rest are from transformer LLMs (sources in Appendix~\Cref{tab:task}). The benchmark includes inference and training kernels, spanning a wide spectrum—from single operators (like Matmul and BatchMatmul) to multi-operator chains (like Matmul+others and LoRA) and larger building blocks (like Group Query Attention and Mamba block). Due to the diversity in the complexity of the baseline kernels, their initial performance also differs a lot from each other, in terms of the achieved percentage of peak throughput. We will also work with the community to enrich the benchmark.

\subsection{Profiling Service}
\label{sec:profiling-service}
\algname requires a profiling service with robust correctness checking and accurate performance measurement to provide reliable feedback signals to maintain an evolving high-quality set of kernels and accumulate useful optimization memory. Due to the vast amount of kernels that need to be sampled to explore the optimization space, it also requires sufficient parallelism in the evolution process.

We check the kernels to be correct under inputs with several different random seeds for $||\mathrm{output} - \mathrm{cpu_{ref}}|| < tol \times ||\mathrm{cpu_{ref}}||$ with a tight $tol$ individually set for each task. For performance measurement, we measure only the execution time, excluding compilation latency. Each round includes warm-up iterations and averages results across multiple runs. To further mitigate fluctuation, we conduct several rounds and select the one with the smallest relative difference, or the first within a predefined threshold. Neuron Profile~\cite{neuron_profile} is used to provide detailed profiling information (full list in Appendix~\Cref{lst:prompt-profile-term}).

\benchname supports \algname\ by efficiently utilizing hardware parallelism.
\algname\ exhibits task-level parallelism because each problem instance runs independently, and sample-level parallelism, where up to $B\times N\times K$ kernels can be profiled simultaneously for each problem. To execute these profiling tasks at scale, the distributed profiling service leverages the core-level and machine-level parallelism of Trainium hardware. Machines are connected via a shared network file system, with a centralized manager dispatching the requests and returning the profiling results. Empirically, cores are periodically rotated to mitigate performance fluctuations after long running.

\begin{figure}[thb]
\begin{center}
\centerline{\includegraphics[width=\columnwidth]{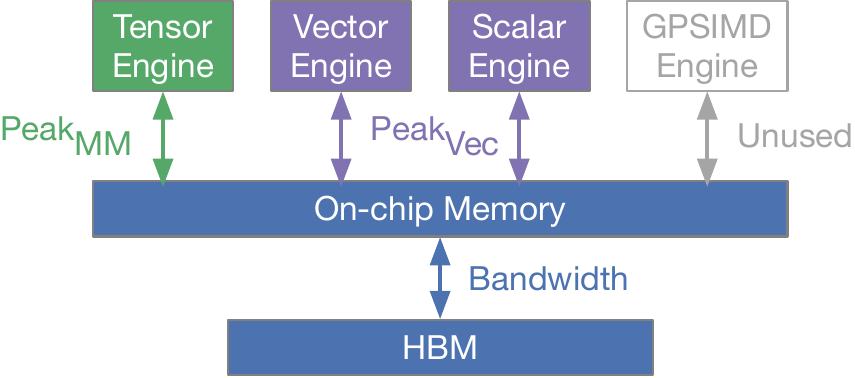}}
\caption{One core of a Trainium chip with its device memory (HBM). Architecture details can refer to NKI docs~\cite{nki_arch}.}
\label{fig:trainium-arch}
\end{center}
\end{figure}

\subsection{Peak Performance Calculation}
Prior work that uses LLMs to write accelerator kernels often measures relative speedup of LLM-generated kernels with respect to some baseline~\cite{ouyang2025kernelbench}, which is an effective metric to demonstrate progress.
For \benchname tasks, we also estimate the best achievable performance offered by the Trainium hardware, which offers additional insights on how effective \algname has been in exploring the entire optimization landscape.

As shown in~\Cref{fig:trainium-arch}, on Trainium chips, tensor, vector, and scalar engines run concurrently and communicate with HBM through kernel-managed on-chip memory. Therefore, using the roofline model analysis~\cite{williams2009roofline}, we calculate the peak performance:
\begin{equation*}
T = \max\!\left(
    \frac{\mathrm{Traffic_{Min}}}{\mathrm{Bandwidth}},
    \frac{\mathrm{FLOPs_{MM}}}{\mathrm{Peak_{MM}}},
    \frac{\mathrm{FLOPs_{Vec}}}{\mathrm{Peak_{Vec}}}
\right)
\end{equation*}
The percentage of peak throughput is calculated as $\frac{T}{t}$, where $t$ is the measured latency. $\mathrm{Traffic_{Min}}$ is the minimal required traffic calculated as the summation of the size of all input tensors and output tensors measured in bytes. We count the matmul FLOPs in Numpy operators as $\mathrm{FLOPs_{MM}}$ and all other FLOPs as $\mathrm{FLOPs_{Vec}}$. We use the summation of peak vector engine and peak scalar engine compute throughput as $\mathrm{Peak_{Vec}}$ because non-matmul instructions can run on these two engines in parallel, and we assume the best case. Hardware specification details are in Appendix~\Cref{tab:peak-perf}. 

Although \algname presents the raw profiling results directly to agents, including the percentage of peak throughput in the prompts could be interesting to investigate.
\begin{figure}[thb]
\begin{center}
\centerline{\includegraphics[width=\columnwidth]{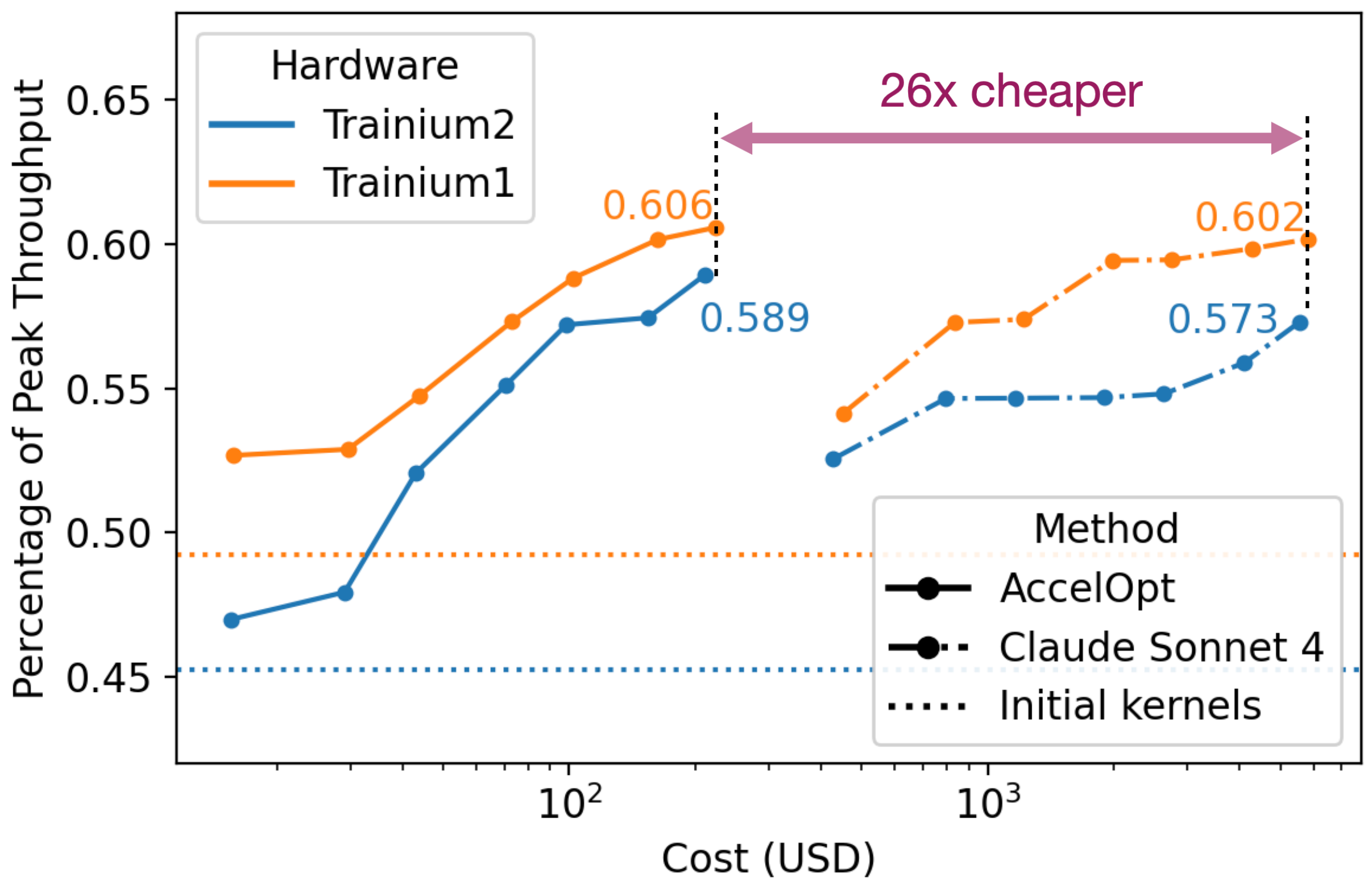}}
\vspace*{-3ex}
\caption{Compare \algname using open-source LLMs with repeated sampling of Claude Sonnet 4 on Trainium 1 and 2.}
\label{fig:main-result}
\end{center}
\end{figure}

\begin{figure*}[thb]
\begin{center}
\centerline{\includegraphics[width=2\columnwidth]{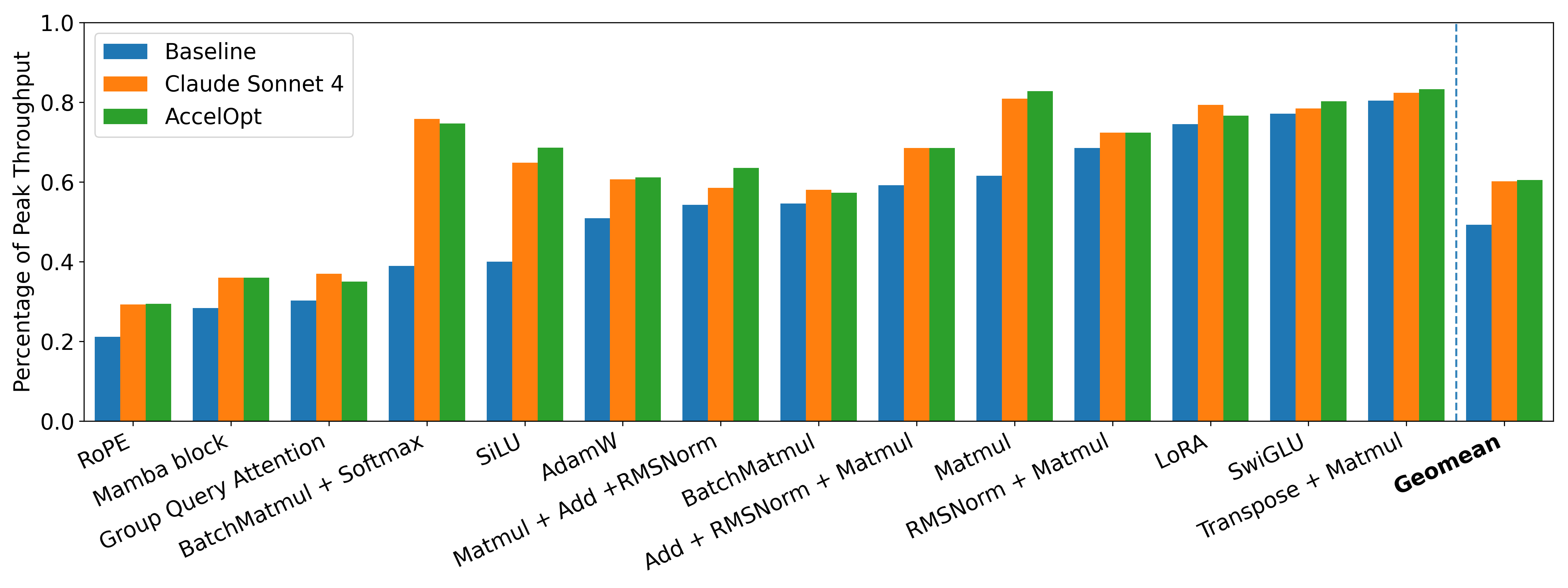}}

\centerline{\includegraphics[width=2\columnwidth]{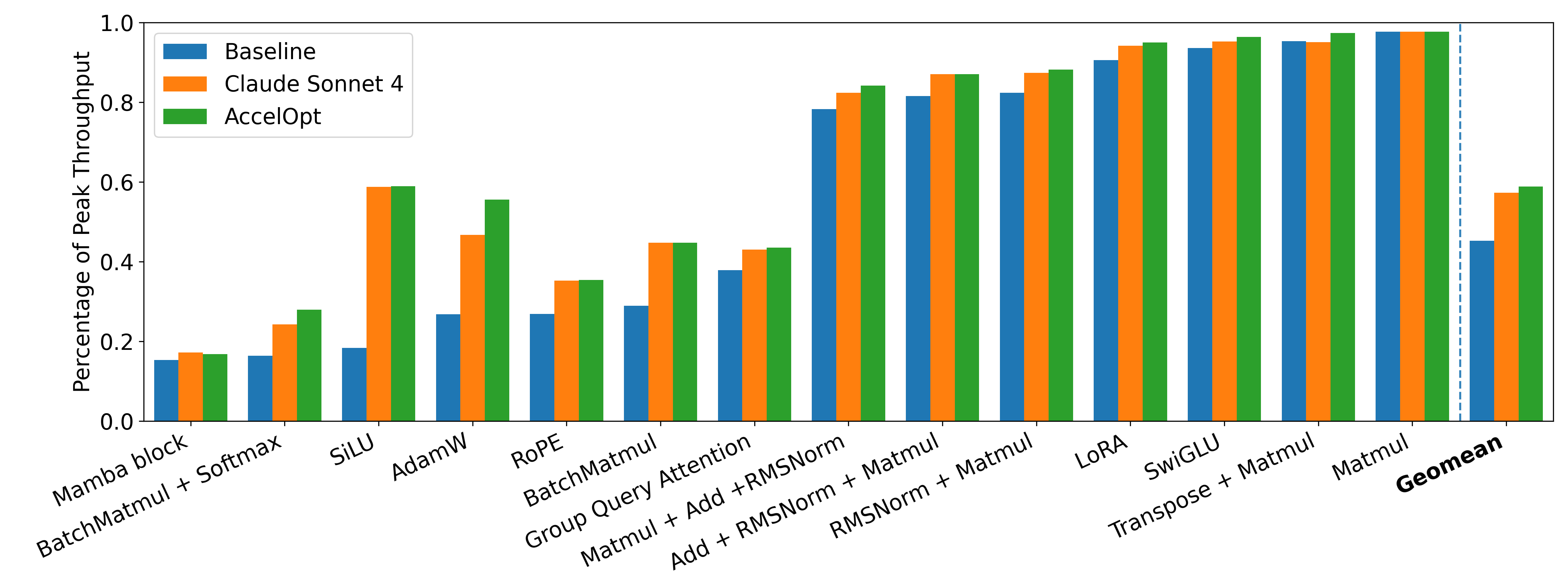}}
\caption{Per-task kernel improvement achieved using Claude Sonnet 4 and \algname on Trainium 1 (above) and Trainium 2 (below). The x-axis is sorted by the baseline kernel's percentage of peak throughput.}
\label{fig:ratio-bars}
\end{center}
\end{figure*}

\section{Evaluation}
\label{sec:evaluation}
We first report the overall performance achieved by \algname (\Cref{sec:overall-perf}). Then, we investigate the optimizations proposed by 
\algname (\Cref{sec:opt-case}) followed by an analysis of its limitations (\Cref{sec:opt-trace}). After that, we conduct an ablation study of the effectiveness of beam search and optimization memory (\Cref{sec:ablation-alg}). Finally, we identify key factors that affect the cost-benefit trade-off (\Cref{sec:cost-analysis}).

\subsection{Overall Performance}
\label{sec:overall-perf}
\textbf{Setup} Claude Sonnet 4 is evaluated using repeated sampling by querying the same prompt multiple times following the test-time scaling practice~\cite{brown2024large}. For \algname, we use Qwen3-Coder-480B-A35B-Instruct-FP8 as the executor model and gpt-oss-120b for the remaining agents with $t_{pos}=1.04$, $t_{neg}=1.15$, TopK=8, ExpN=16, B=6, N=12, and T=16. The prompt for Claude Sonnet 4 is in Appendix \Cref{lst:claude-prompt-a,lst:claude-prompt-b,lst:claude-prompt-c}, similar to that for \algname.

\textbf{Performance} \Cref{fig:ratio-bars} shows the achieved percentage of peak throughput on Trainium 1 and Trainium 2, where \algname performs comparably with Claude Sonnet 4 across most kernels. As shown in \Cref{fig:main-result}, \algname improves the average throughput from $49\%$ to $61\%$ of peak on Trainium 1 and from $45\%$ to $59\%$ on Trainium 2—matching Claude Sonnet 4 (thinking mode) while being $26\times$ cheaper. 
Moreover, using Claude Sonnet 4 as the base model in \algname can reduce expenses by $3.3\times$ compared with repeated sampling as shown in~\Cref{tab:planner-executor-tune}. 
Since Claude Sonnet 4’s internal reasoning tokens are unavailable, cost is defined as the sum of input and output tokens multiplied by the per-token price listed in the Appendix \Cref{tab:token-cost}.
\begin{figure*}[thb]
\begin{center}
\centerline{\includegraphics[width=2\columnwidth]{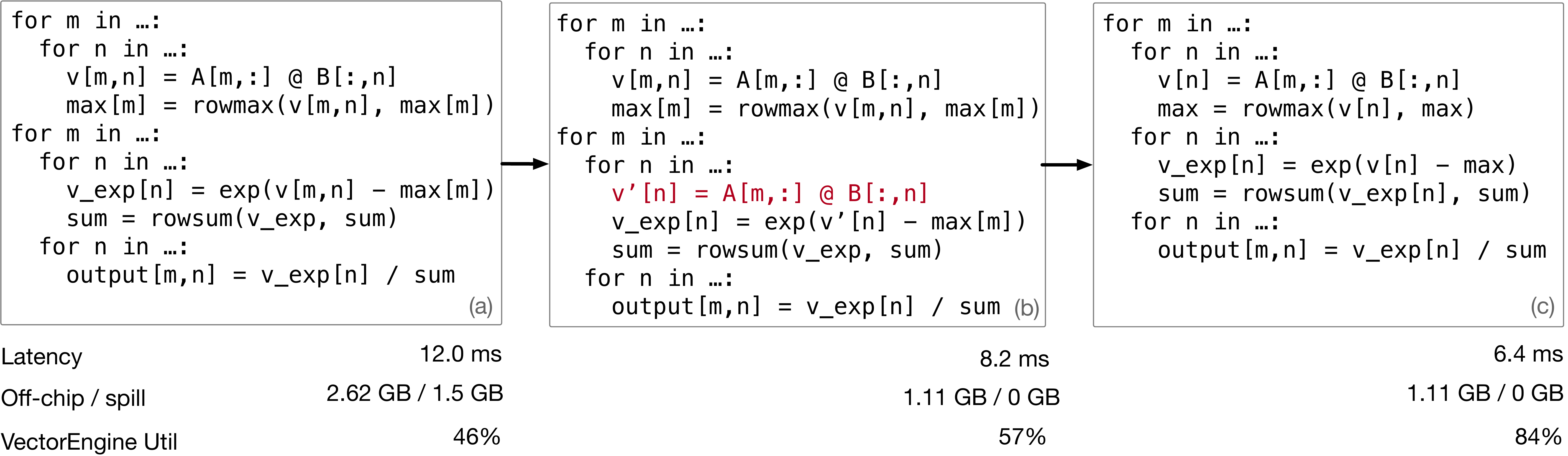}}
\vspace*{-2ex}
\caption{Non-local optimization discovered by \algname for the fused BatchMatmul+Softmax operator. All variables are tiles of tensors, and code has been simplified to highlight the changed dimensions of allocated tensors in the loop body.}
\label{fig:kernel-case-study}
\end{center}
\end{figure*}

\subsection{Optimization Case Study}
\label{sec:opt-case}
We exemplify a few cases of intriguing optimizations discovered using \algname to illustrate its strengths.

\paragraph{Peephole Optimization}
\algname can accomplish peephole optimizations like algebraic simplification and hardware-level intrinsic fusion. For example, \algname simplifies the expression $\theta_{t-1} - \gamma \lambda \theta_{t-1}$ to $(1 - \gamma \lambda)\theta_{t-1}$, enabling precomputation of $(1 - \gamma \lambda)$. Additionally, \algname can recognize idiomatic instruction patterns such as
\texttt{reciprocal(sqrt(...)))} $\Rightarrow$ \texttt{rsqrt(...))},
which reduces intermediate tensors.
For SiLU, \algname can conduct transformation $x/(1 + e^{-x}) \Rightarrow x \cdot \text{sigmoid}(x)$ to leverage NKI’s specialized instruction, resulting in more efficient execution.

\paragraph{Loop Optimization}
Apart from peephole optimizations, \algname can also discover non-local optimizations such as loop transformations. We pick two snapshots from one optimization trace of BatchMatmul + Softmax as shown in~\Cref{fig:kernel-case-study}. The baseline kernel (a) results in memory spilling because tiles \texttt{v} and \texttt{p} have to live across two loops. LLM agents identify this inefficiency and manage to remove the spilling by recomputing \texttt{v'} at kernel (b). Although this optimization reduces off-chip memory access, which improves the performance, it also introduces an extra matrix multiplication before the \texttt{exp}. 
Therefore, LLM agents decide to remove the recomputation and the extra \texttt{m} loop in kernel (c). In this way, the generated kernel achieves no spilling and higher vector engine utilization. This indicates that \algname is capable of discovering global optimizations that require multiple steps of non-trivial reasoning that involve kernel semantics, the underlying hardware architecture, and profiler feedback.

\paragraph{Educational Impact}
In Stanford CS149 Fall 2025, a graduate-level parallel computing course, we used \algname to optimize a Conv2D kernel outside of \benchname and achieved 48.8\% of peak throughput starting from last year's reference implementation (9.54\%). Based on the optimization proposed by \algname, we designed an extra credit problem. 33.6\% of 131 teams of students successfully conquered the challenge. Throughout the process, the students learned two principles: (1) transforming sequential temporal iteration to parallel spatial execution, and (2) specialization for workload under hardware constraints. This example underscores the generality of \algname beyond \benchname and highlights the educational impact of LLM-assisted kernel optimization.

\paragraph{Comparison with Human Experts} To quantify how close \algname comes to expert-level results, we tracked its progress on two kernels with human-optimized reference versions. 
\textbf{(1) Mamba.}
The NKI tutorial~\cite{nki_tutorial}
provides three progressively faster human versions, reaching $28.4\%$, $30.1\%$, and $52.7\%$ of peak throughput.
Starting from the same baseline ($28.4\%$ of peak), \algname autonomously improved the kernel to $54.6\%$ of peak, which is $1.04\times$ the best expert result ($52.7\%$).
Moreover, the generated kernel used a different loop order than the best human. \textbf{(2) RoPE.}
The initial RoPE kernel was adopted from nki-samples~\cite{nki_samples}, which provides one version of RoPE ($21.1\%$ of peak).
Starting from this version, \algname improved performance to $29.6\%$ of peak, a $1.4\times$ speedup over the human reference. These results demonstrate that AccelOpt can exceed expert-level performance. This stems from AccelOpt's scalability: human experts optimize a handful of kernels sequentially, while AccelOpt can explore many in parallel.

\subsection{Optimization Limitation Analysis}
\label{sec:opt-trace}
In~\Cref{sec:opt-case}, we investigate what \algname can achieve. To understand the limitations of \algname, we analyze the saturating behaviors where \algname cannot further optimize the kernels with more iterations.

We observed two causes of saturating behaviors: (1) \algname can still do effective exploration but cannot further improve the kernel performance because the kernel is close to peak, (2) \algname cannot do effective exploration because the initial kernel is challenging to optimize.

 We use the trend and variation of performance metrics as a proxy for the effectiveness of optimization space exploration. Traffic efficiency measures how much of the data movement is necessary, defined as:
\begin{equation*}
\mathrm{Traffic Efficiency} = \frac{\mathrm{Traffic_{Min}}}{\mathrm{HBM_{Read}} + \mathrm{HBM_{Write}}}
\end{equation*}
Engine utilization (tensor, vector, or scalar) is directly obtained from profiling, which measures the ratio of engine active time to total kernel execution time, reflecting how busy the engine is. All metrics range from 0 to 1.

\begin{figure}[htb]
\centering
\includegraphics[width=\linewidth]{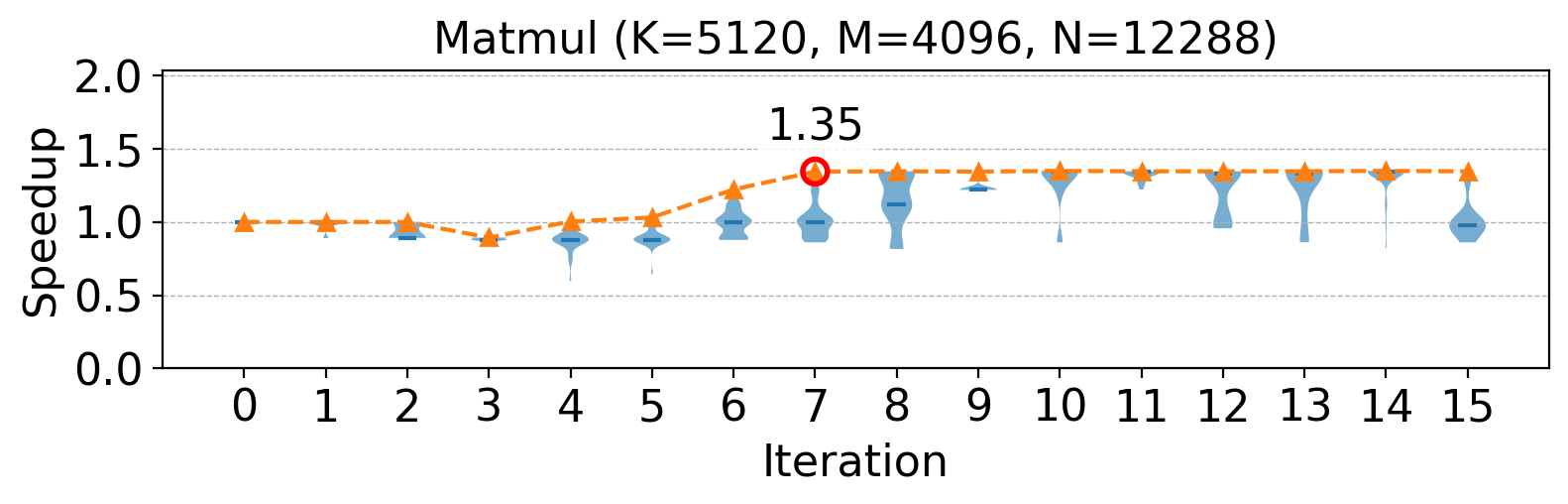}
\includegraphics[width=\linewidth]{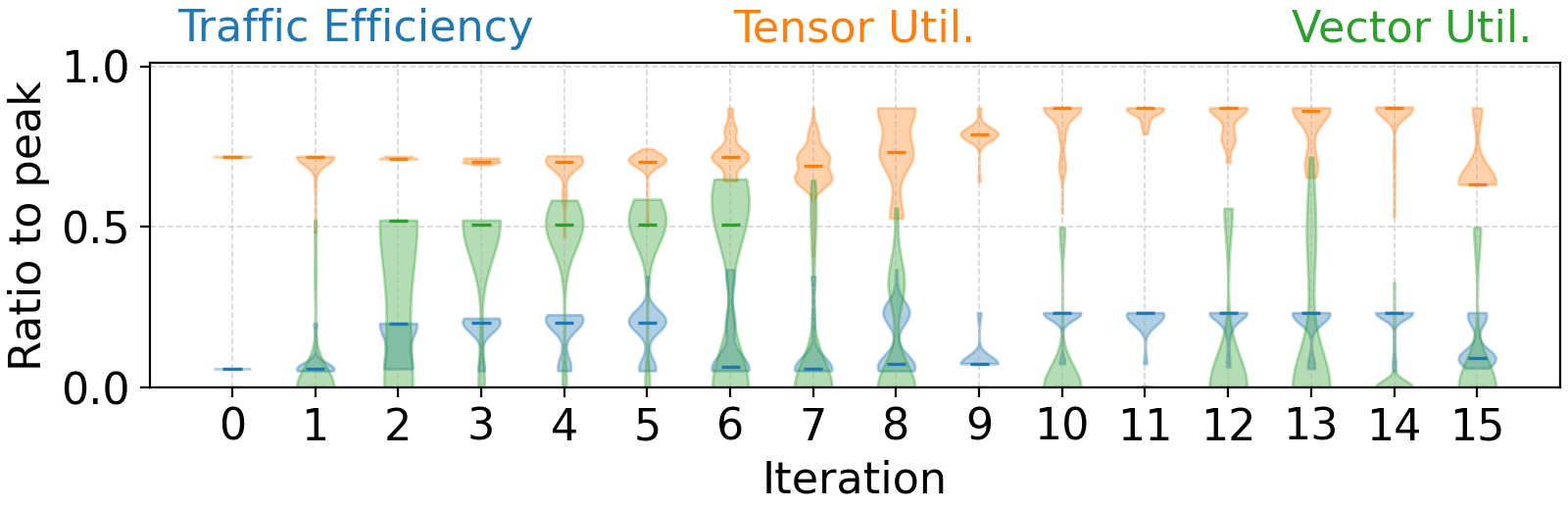}
\vspace*{-3ex}
\caption{Saturating speedup with effective exploration. Same as~\Cref{fig:case-transpose-matmul} and~\Cref{fig:case-bmm}, the above panel is the speedup distribution of all generated kernels; the below is the distribution of additional performance metrics at each iteration.}
\label{fig:case-matmul}
\end{figure}

As shown in~\Cref{fig:case-matmul}, although the case exhibits a growing speedup window, the maximum speedup plateaus after iteration~7. This plateau does not indicate that the agent is merely repeating previous successful experiences, a pitfall when using self-generated in-context examples in optimization~\cite{wan2025few}. 
Instead, the agent continues to explore diverse strategies that visibly affect performance: at iteration~10, the traffic efficiency shifts to a new distribution, and the maximum vector utilization continues to vary beyond iteration~7. Therefore, the exploration mechanism remains active; speedup saturates because the kernel discovered at iteration~7 has already reached about 82\% of peak throughput, leaving little room for further improvement.

\begin{figure}[htb]
\centering
\includegraphics[width=\linewidth]{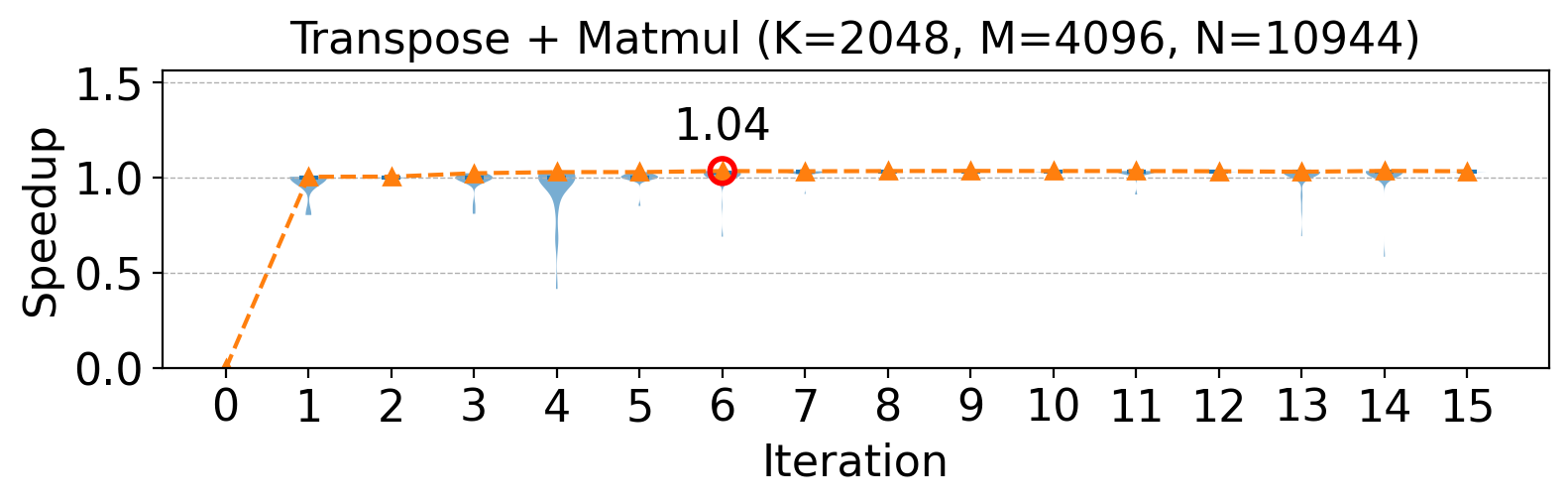}
\includegraphics[width=\linewidth]{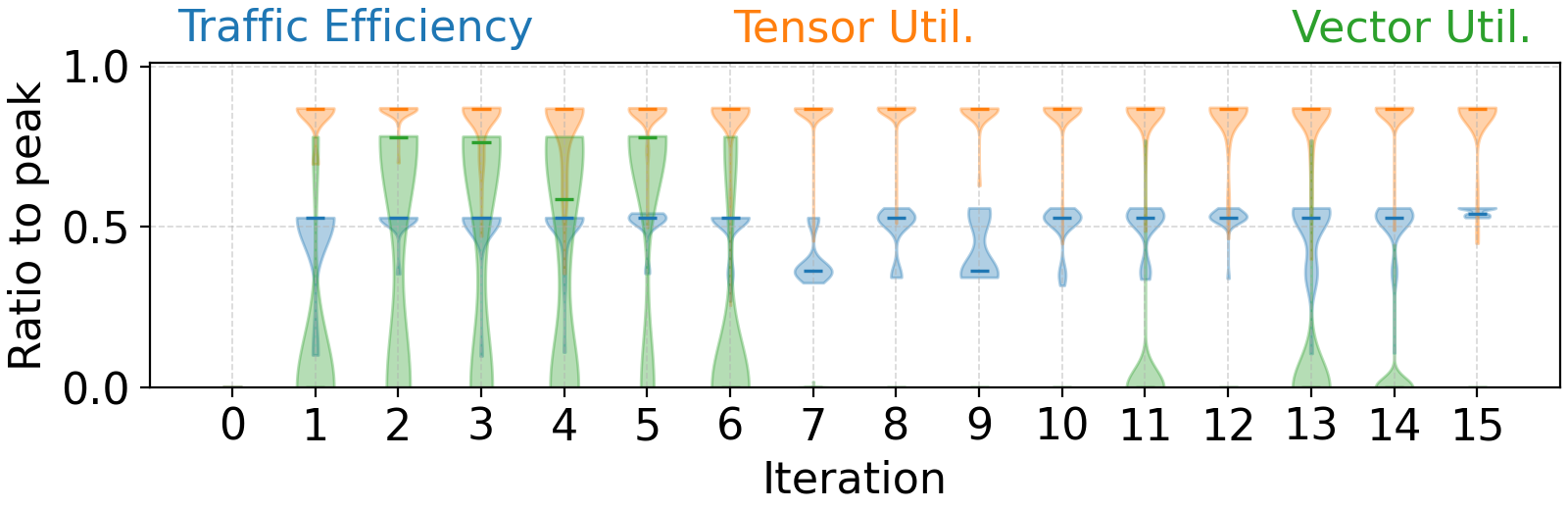}
\vspace*{-3ex}
\caption{Early saturating speedup with effective exploration. The non-reduction dimension N is large, leading to a wide memory-management search space. However, because the operator is dominated by matrix multiplication and the baseline already reaches about 83\% of peak throughput, further gains are difficult.}
\label{fig:case-transpose-matmul}
\end{figure}

Unlike~\Cref{fig:case-matmul}, the speedup in~\Cref{fig:case-transpose-matmul} saturates early. Although latency does not improve, the agents continue to propose meaningful rewrites rather than minor local tweaks. Their exploration is reflected in the large variations and shifting trends in vector utilization and traffic efficiency.

\begin{figure}[htb]
\centering
\includegraphics[width=\linewidth]{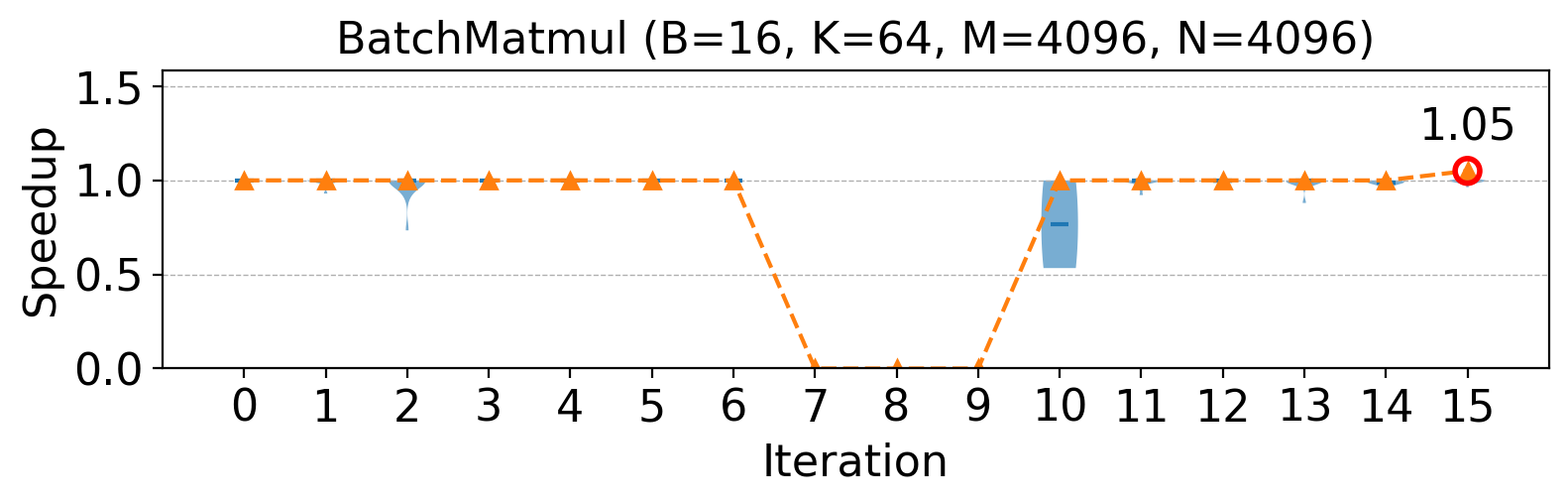}
\includegraphics[width=\linewidth]{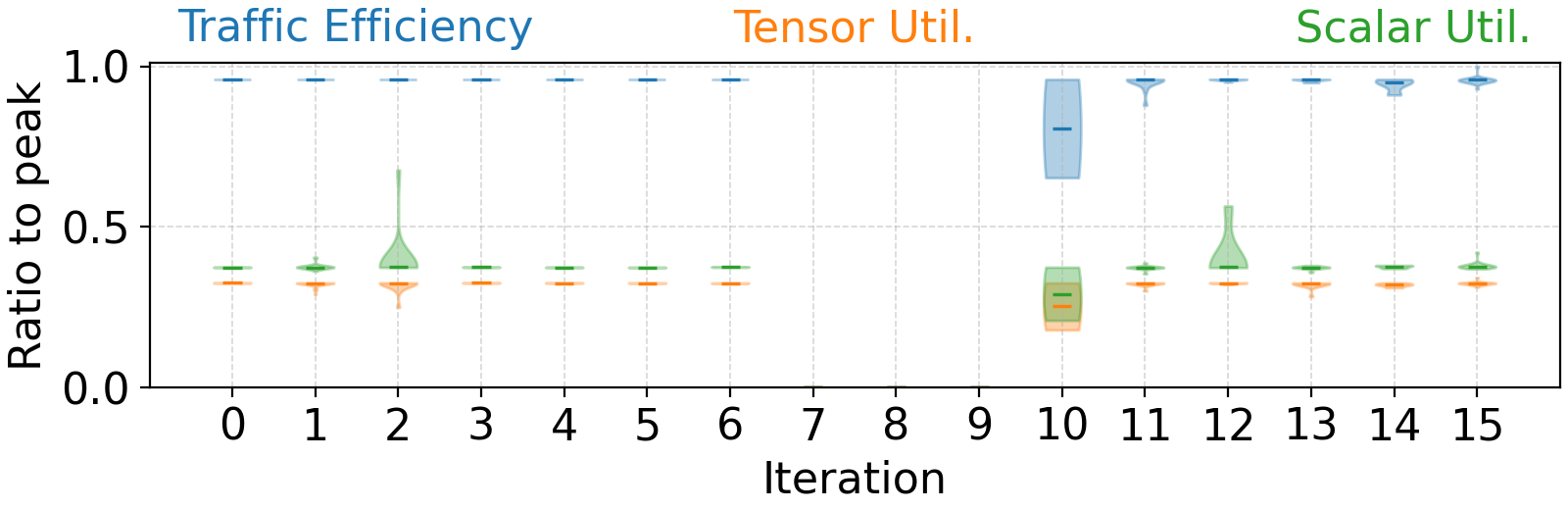}
\vspace*{-4ex}
\caption{Saturating speedup without effective exploration. Vector engine utilization is nearly zero, thus we plot scalar engine utilization here. }
\label{fig:case-bmm}
\end{figure}

In the case of~\Cref{fig:case-bmm}, very few effective rewrites are discovered by \algname. All the performance metrics barely change, and at iterations 7-9, no correct kernels are generated.  This is because the problem is challenging to optimize: the problem size is small enough for all the data to fit on-chip, which causes the traffic efficiency to be nearly 100\% in the baseline, and the reduction dimension K=64 is half of the hardware-native reduction dimension (128), so it is hard to fully utilize the tensor engine using current NKI APIs.

To be noted, we post failure cases in this section, but there are actually runs where diverse exploration leads to post-plateau improvements as shown in Appendix~\Cref{fig:case-bmm-softmax}. 

\subsection{Ablation Study of \algname Components}
\label{sec:ablation-alg}
\begin{figure}[thb]
\centering
\includegraphics[width=\linewidth]{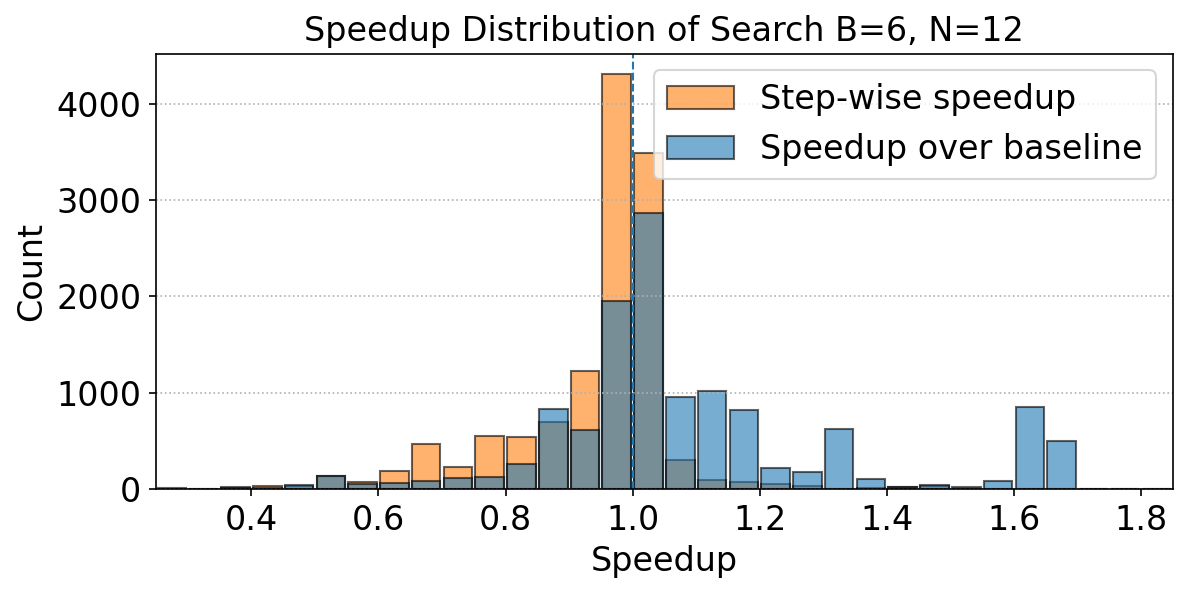}
\vspace*{-5ex}
\caption{The orange bars show distribution of per-iteration speedup over the candidate kernels, while the blue bars show the speedup over the initial kernels. This plot collects distribution of speedups from all tasks.}
\label{fig:speedup-dist}
\end{figure}
\begin{figure}[thb]
\begin{center}
\centerline{\includegraphics[width=\columnwidth]{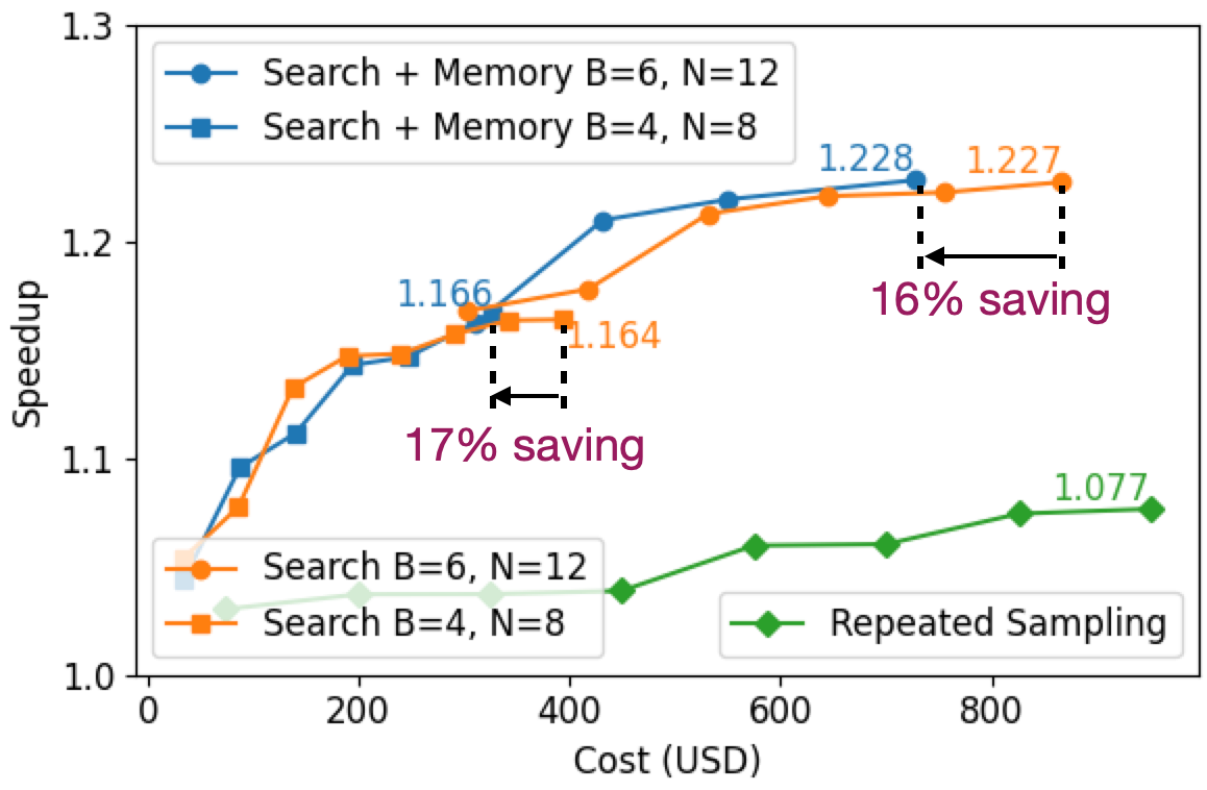}}
\vspace*{-2ex}
\caption{Geometric mean of best speedup achieved up to a certain iteration across all tasks obtained through repeated sampling, beam search, and beam search + optimization memory. As defined in~\Cref{alg:nkiopt}, $B$ is the number of candidates and $N$ is the number of plans for each candidate. }
\label{fig:sweep-params}
\end{center}
\end{figure}

\paragraph{Beam Search vs. Repeated Sampling Only}
As shown in~\Cref{fig:sweep-params}, beam search outperforms repeated sampling of the agentic workflow, using the same LLMs. This is because each iteration builds upon previous best kernels, leading to progressively better optimizations. In~\Cref{fig:speedup-dist}, the orange bars cluster near $1.0\times$, whereas the blue bars include more cases exceeding $1.0\times$, confirming that beam search yields cumulative performance gains.

\begin{figure}[htb]
\centering
\includegraphics[width=\linewidth]{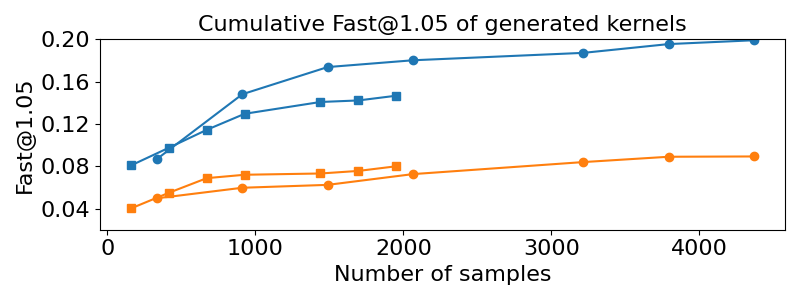}\\[0.6em]
\includegraphics[width=\linewidth]{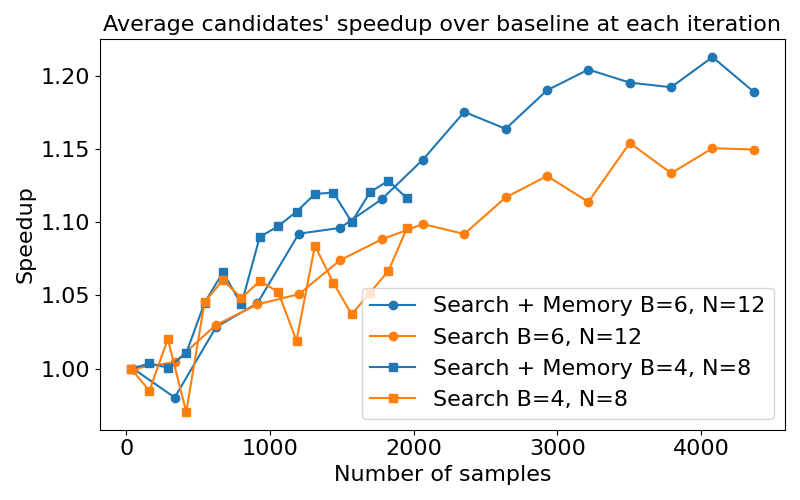}
\vspace*{-5ex}
\caption{Optimization memory improves cost-efficiency, leading to a higher percentage of good performing kernels (top) by improving the kernel quality per-iteration (below). 
To compare systems with different search hyperparameters, we consider the number of kernels sampled rather than number of iterations. 
 Note that the average current iteration candidates' speedup over baseline can drop below $1.0\times$ because $\beta$ selects from all correct kernels, not only those with speedups.}
\label{fig:speedup-candidates}
\end{figure}

\paragraph{Optimization Memory vs. Beam Search Only}
As shown in~\Cref{fig:sweep-params}, search-only experiments run the total T=16 iterations, while Search + Memory experiments achieve similar speedup in 13 iterations, saving 16-17\% cost. Optimization memory increases the probability of generating fast kernels (higher cumulative Fast@p~\cite{ouyang2025kernelbench}), yielding stronger candidate pools and, ultimately, higher best speedups using fewer iterations (see \Cref{fig:speedup-candidates}).
The candidate speedup is the geometric mean of the candidate kernels' speedup over the initial kernel. We use cumulative Fast@p defined as:
\begin{equation*}
\mathrm{Fast}@p = \frac{1}{N} \sum_{i=1}^{N} \mathbb{I}(\text{correct}_i \land \{\text{speedup}_i > p\})
\end{equation*}
where $N$ refers to all generated kernels until the current iteration. Based on the comparisons in~\Cref{fig:sweep-params}, we use B=6 and K=12 for other experiments.

\paragraph{Validate the Diversity Design}
To justify AccelOpt’s design for diversity, we conduct experiments with B=1, matching the number of profiled kernels with~\Cref{tab:executor-tune} gpt-oss-120b by setting N=72 and keeping K=2, T=16. B=1 with memory achieves final best speedup 1.204$\times$ (\$143.58), and B=1 without memory achieves 1.229$\times$ (\$116.35). Both are worse than 1.235$\times$. Comparing 1.229$\times$ with 1.204$\times$ implies that the memory might limit optimization if the candidates are not diverse enough. When B and K are small, the best speedup drops as shown by the comparison between (B=4, N=8) and (B=6, N=12) in~\Cref{fig:sweep-params}. Therefore, we expect the performance to further drop when decreasing B and N to (1,2,4).

\paragraph{Reflexion-style Baseline}
We also implemented a Reflexion-style workflow where a reflector generates optimization insights at each iteration, managed following the Reflexion paper~\cite{shinn2023reflexion}. We match the number of generated kernels with~\Cref{tab:executor-tune} gpt-oss-120b by launching 144 (B(=6)$\times$ N(=12)$\times$ K(=2)) trajectories each with T=16 iterations. The Reflexion-style baseline achieves 1.137$\times$ speedup (\$178.37) versus 1.235$\times$ (\$139.00) for \algname. More tokens are consumed because the Reflexion-style baseline reflects on every generated kernel, while \algname reflects on a selected group of kernels per iteration. This demonstrates the effectiveness of AccelOpt's memory mechanism. 

\paragraph{Non-LLM Search-based Baselines}
Our goal is fully autonomous kernel optimization without manual optimization insight curation/ingestion, whereas non-LLM search-based optimizers typically rely on substantial manual effort and deep architecture knowledge, including rewriting initial kernels to match the optimizer’s input format and explicitly specifying a search space together with pruning strategy as in~\cite{wu2025mirage}. It is thus difficult to set up a fair comparison with these methods. 

\subsection{Cost Analysis}
\label{sec:cost-analysis}
This section conducts cost analysis to identify key factors in optimization memory configuratin and best models that affect the cost-benefit trade-off. The benefit is meansured by the geometric mean across all problems of each problem’s maximum speedup achieved over all iterations.

\begin{figure}[htb]
\centering
\includegraphics[width=0.95\linewidth]{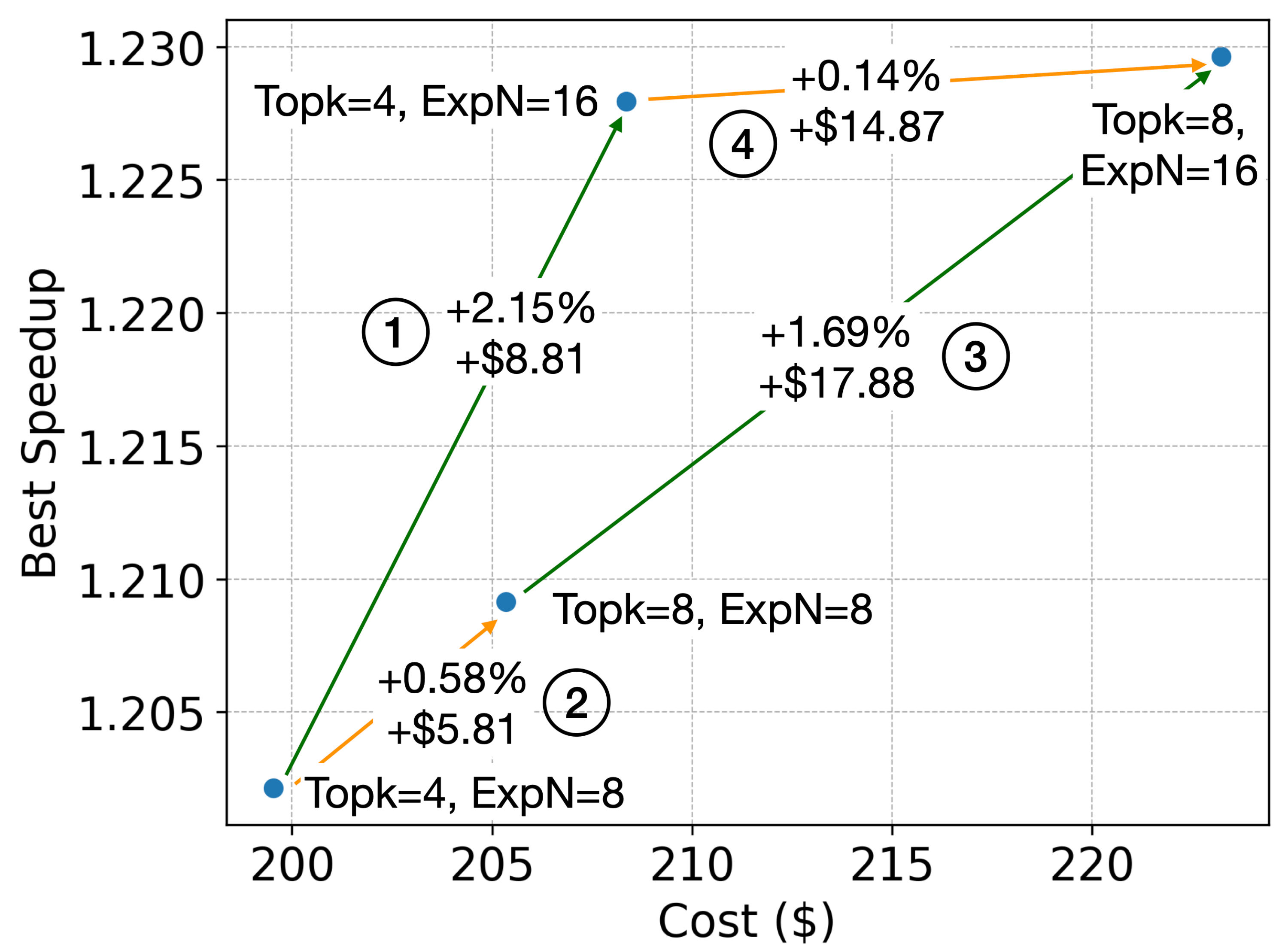}
\vspace*{-2ex}
\caption{Cost-benefit trade-off across different TopK and ExpN. As the ExpN increases, $\sigma$ can collect experiences from more iterations preceding the last iteration. As the TopK increases, $\sigma$ can collect more experiences from the current iteration.}
\label{fig:window-sweep}
\end{figure}

\textbf{Increasing memory capacity (ExpN) is more cost-efficient than increasing memory update eagerness (TopK)}. As shown in~\Cref{fig:window-sweep}, increasing TopK and ExpN both can help increase the best speedup. Comparing $\text{\textcircled{\footnotesize 1}}$ and $\text{\textcircled{\footnotesize 3}}$  with $\text{\textcircled{\footnotesize 2}}$ and $\text{\textcircled{\footnotesize 4}}$, under similar cost, the delta of speedup is much larger when increasing ExpN than when increasing the TopK. 
Therefore, we use TopK=8, ExpN=16 for experiments in~\Cref{sec:overall-perf}.

We also observe that the effectiveness of increasing ExpN depends on the model. As shown in~\Cref{tab:executor-tune}, Qwen3-Coder-30B gains 4.6\% speedup improvement with extra \$12.33 cost. On the contrary, the extra \$13.81 cost only brings 0.6\% speedup improvement in gpt-oss-120b.  

\textbf{Switching base models for agents can have different cost-benefit trade-offs}. As shown in~\Cref{tab:executor-tune}, the executor model needs to be capable enough to understand and correctly implement the plan. Qwen3-Coder-30B and Qwen3-Coder-480B come from the same model family, and the larger one gets better performance. Qwen3-Coder-30B and gpt-oss-120b have the same cost per token, while gpt-oss-120b is a reasoning model. The extra reasoning tokens buy a higher speedup. Allowing different models to serve as the executor could further improve performance. This model ensemble strategy selects the best kernel for each problem from the experiments run by each executor. As shown in the last row of~\Cref{tab:executor-tune}, the model ensemble achieves the best results for both ExpN = 8 and ExpN = 16, with total cost equal to the sum of the costs of the three single executor experiments.

\begin{table}[htb]
\caption{Best speedup and cost comparison across different ExpN and executor model settings, fixing gpt-oss-120b as planner and summarizer.}
\label{tab:executor-tune}
\vskip 1ex
\centering
\small
\begin{tabular}{c c c c}
\toprule
\diagbox{Executor}{ExpN} & 8 & 16 & Delta \\
\midrule
Qwen3-Coder-30B- & 1.144 & 1.197 & +4.6\% \\
A3B-Instruct & \$96.10 & \$108.43 & +\$12.33 \\
\midrule
gpt-oss-120b & 1.228 & 1.235 & +0.6\% \\
& \$125.19 & \$139.00 & +\$13.81 \\
\midrule
Qwen3-Coder-480B- & 1.209 & 1.230 & +1.7\% \\
A35B-Instruct-FP8& \$205.35 & \$223.23 & +\$17.88 \\
\midrule
Ensemble & 1.241 & \textbf{1.246} & +0.4\% \\
(Best of three models)& \$426.64 & \$470.66 & +\$44.02 \\
\bottomrule
\end{tabular}
\end{table}

As shown in~\Cref{tab:planner-tune}, different from switching executors, we did not observe substantial differences in speedup when switching planners. This implies that \textbf{further performance improvements could first focus on enhancing the executor’s capability.}

\begin{table}[htb]
\caption{Switching planners' base models using the best configuration discovered from~\Cref{fig:window-sweep} and~\Cref{tab:executor-tune}: gpt-oss-120b as executor with Topk=8, ExpN=16.}
\label{tab:planner-tune}
\vskip 1ex
\centering
\small
\begin{tabular}{c c c}
\toprule
Planner & Speedup & Cost \\
\midrule
gpt-oss-20b & 1.234 & \$116.87 \\
\midrule
gpt-oss-120b & 1.235 & \$139.00 \\
\midrule
Qwen3-235B-Thinking & 1.234 & \$316.21 \\
\bottomrule
\end{tabular}
\end{table}

Using Claude Sonnet 4 as agent backbones in \algname can reduce expenses compared with repeated sampling, but is still more expensive than only using open-source models. \Cref{tab:planner-executor-tune} indicates that using gpt-oss-120b for both the planner and executor achieves the best performance and cost. Although extensive hyperparameter tuning was limited due to cost constraints, \algname already reduces expenses compared with repeated sampling speedup 1.222$\times$ (\$5806.83).
\begin{table}[H]
\caption{Apply Claude Sonnet 4 to \algname.}
\label{tab:planner-executor-tune}
\vskip 1ex
\small
\centering
\begin{tabular}{c c c}
\toprule
Planner & Executor & Speedup (Cost) \\
\midrule
Claude Sonnet 4 & Claude Sonnet 4 & 1.226 (\$1732.73) \\
\midrule
gpt-oss-120b & Claude Sonnet 4 & 1.213 (\$1269.98) \\
\midrule
Claude Sonnet 4 & gpt-oss-120b & 1.208 (\$1223.05) \\
\midrule
gpt-oss-120b & gpt-oss-120b & 1.235 (\$139.00) \\
\bottomrule
\end{tabular}
\end{table}

\subsection{Generalization to Other Platforms}
\algname is platform-agnostic: the beam search and self-improving memory mechanism are orthogonal to specific hardware. Adaptation requires: (1) a profiling service and (2) platform-specific base prompts. \algname can also apply the same methodology to communication operations, but this paper only focuses on single-core kernels without cross-chip communication. Extending to communication primitives is valuable for full-stack performance and represents a promising direction for future work. Critically, our Trainium evaluation demonstrated that AccelOpt's evolution approach works when LLMs have limited prior knowledge of the platform. Thus, the technique should be even more effective on mature platforms like GPUs, where LLMs have more relevant training data. On 24 Triton kernels from FlashInfer-Bench~\cite{fib} (H100), AccelOpt with gpt-oss-120b achieved $1.27\times$ average speedup over best Triton baselines, with $3.19\times$ peak speedup on a GQA decoding kernel. The complete results are in Appendix~\Cref{fig:fib-results}. 

\section{Related work}

\textbf{Memory for LLM Agents.} Memorizing past experiences has been shown to be critical for developing self-evolving agent systems~\cite{zhang2025agentic,sun2025seagent,ouyang2025reasoningbank}.
LessonL~\cite{liu2025lessons} has explored evolving memory in the context of CPU code optimization. LessonL’s memory is composed of experience items whose performance anchor is always the baseline kernel. Therefore, their mechanism may be prone to homogeneity, which is mitigated by LLM ensembling (c.f. Table 10 in~\citet{liu2025lessons}). On the contrary, AccelOpt’s memory is evolving with the candidate kernels and thus could be more diverse. 

\textbf{LLM Agents for AI Accelerator Kernel Optimization.}
The agentic system proposed by~\citet{zhang2025adaptive} translates ML operators to AI accelerator kernels but cannot optimize them.
Autocomp~\cite{hong2025autocomp} optimizes unfused kernels, and its planners rely on manually crafted, problem-specific lists of optimizations. AlphaEvolve~\cite{novikov2025alphaevolve} optimizes matrix multiplication and FlashAttention kernels on TPUs, but the system implementation is not publicly available.
GEPA~\cite{agrawal2025gepa} improves LLM-generated AMD NPU kernels by evolving prompts through automatic discovery and injection of architectural best practices on NPUs. This method can potentially be used to produce better prompts for the executor agent in \algname.
However, the optimization memories discovered by \algname can potentially provide more detailed task-specific insights (c.f. Figure~25 in \citet{agrawal2025gepa} vs. \Cref{lst:output-summarizer-a,lst:output-summarizer-b,lst:output-summarizer-c} in Appendix).

\textbf{Benchmarks for Kernel Optimization.}
Various benchmarks have been proposed for kernel optimization on GPUs and AI accelerators~\cite{ouyang2025kernelbench,wen2025multikernelbench,tian2025heterobench}. Recent work also improve interface usability~\cite{saroufim2025backendbench,fib} and evaluation robustness~\cite{lange2025towards,zhangkernelbot}. These benchmarks usually measure relative kernel speedup compared to certain performance baselines. Yet \benchname also measures kernel performance using the ratio to peak throughput, offering an absolute standard to understand kernel performance on a given hardware platform.

\section{Conclusion}
This paper presents \algname, the first self-improving LLM agentic system for kernel optimization on emerging AI accelerators that combines search with memory accumulation. 
We demonstrate that combining inference-time scaling with optimization memory enables LLM agents to autonomously optimize real-world Trainium kernels in \benchname, our curated benchmark suite, without requiring expert optimization knowledge. 
Through systematic ablation studies, we confirm the effectiveness of beam search and optimization memory in obtaining high-performing kernels with improved cost efficiency. 
We also find that open-source models achieve higher cost efficiency than leading proprietary coding models for this task. 
Overall, \algname and \benchname provide a promising foundation for automated kernel optimization on emerging AI accelerators.

\section*{Acknowledgment}
We are grateful to anonymous reviewers and the shepherd. We also thank the Amazon Neuron Science team, Stanford Pervasive Parallelism Lab, Stanford CS149 Fall 2025 team, FlashInfer-Bench team, Tian Zhao, Simon Guo, Anne Ouyang, Yuhui Zhang, Ching-An Cheng, and many others for the helpful discussions and feedback throughout this project. 
This work was supported in part by DARPA under the Machine learning and Optimization-guided Compilers for Heterogeneous Architectures (MOCHA) program (award number HR00112520038). Any opinions, findings, and conclusions or recommendations expressed in this material are those of the authors and do not necessarily reflect the views of the aforementioned funding agencies.
\bibliography{reference}
\bibliographystyle{mlsys2025}

%%%%%%%%%%%%%%%%%%%%%%%%%%%%%%%%%%%%%%%%%%%%%%%%%%%%%%%%%%%%%%%%%%%%%%%%%%%%%%%
%%%%%%%%%%%%%%%%%%%%%%%%%%%%%%%%%%%%%%%%%%%%%%%%%%%%%%%%%%%%%%%%%%%%%%%%%%%%%%%
% SUPPLEMENTAL CONTENT AS APPENDIX AFTER REFERENCES
%%%%%%%%%%%%%%%%%%%%%%%%%%%%%%%%%%%%%%%%%%%%%%%%%%%%%%%%%%%%%%%%%%%%%%%%%%%%%%%
%%%%%%%%%%%%%%%%%%%%%%%%%%%%%%%%%%%%%%%%%%%%%%%%%%%%%%%%%%%%%%%%%%%%%%%%%%%%%%%
\clearpage
\appendix

\section{Appendix}
\subsection{Extra Information}
The compilation time does not affect kernel quality since kernels are usually reused multiple times in ML pipelines after one-time compilation. We measured compilation time for the best kernels from one \algname run: \algname solutions took 1.59–31.29s (mean 7.38s), while baseline \benchname kernels took 1.77–30.61s (mean 8.13s).
\begin{figure}[htbp]
\centering
\includegraphics[width=\linewidth]{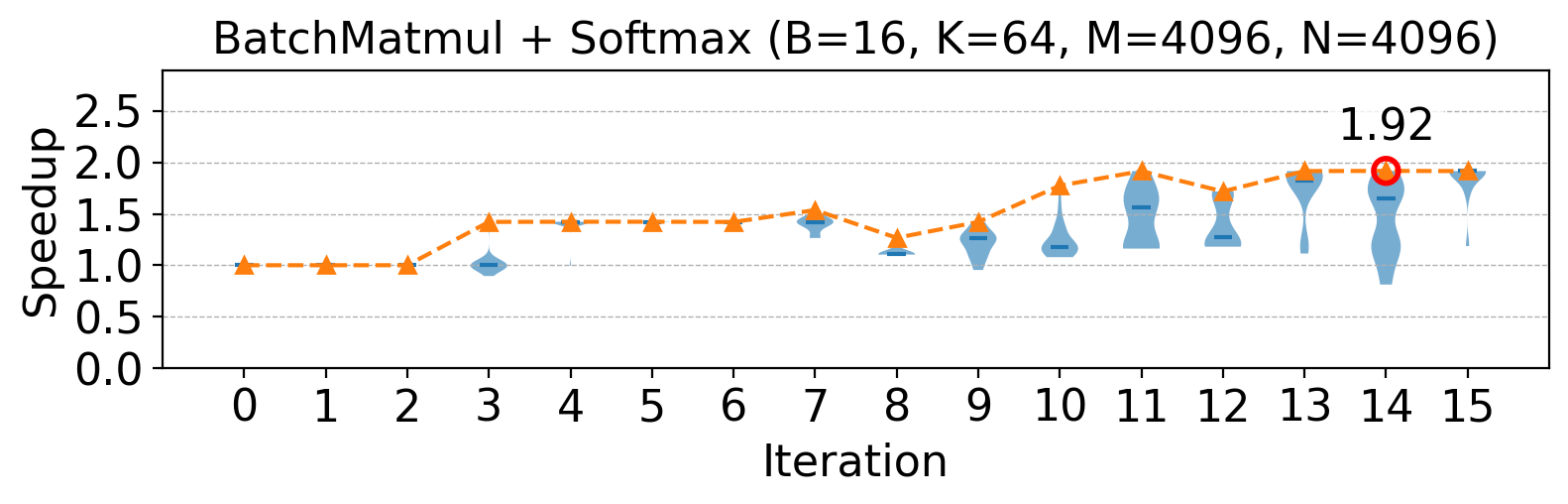}
\includegraphics[width=\linewidth]{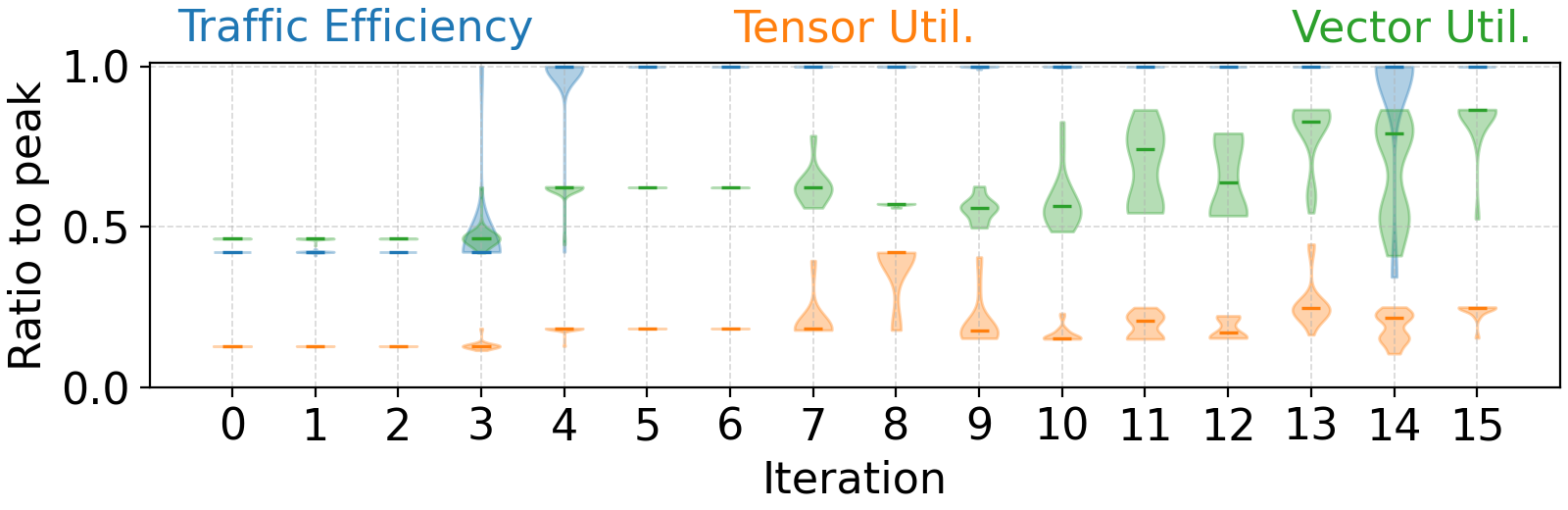}
\caption{An example showing diverse exploration leads to post-plateau improvements. We observed a run where the best speedup plateaued from iterations 3 to 8 (and even degraded), but continued exploration still changed key utilization metrics; from iteration 9 onward, non-trivial improvement began to emerge, and the best kernel improved again until iteration 14.}
\label{fig:case-bmm-softmax}
\end{figure}

\begin{figure}[ht]
    \centering
    \includegraphics[width=1.0\linewidth]{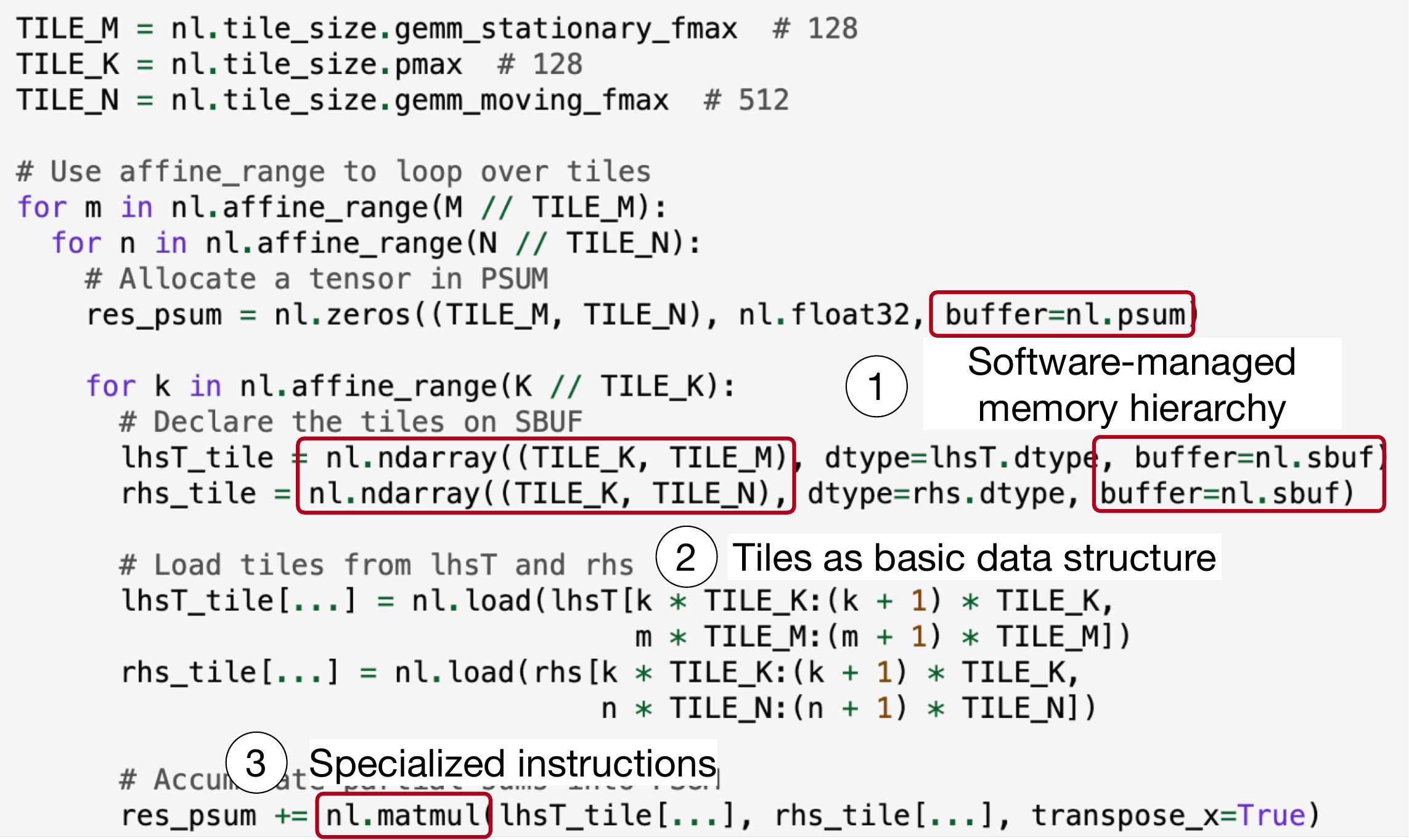}
    \caption{An example NKI program snippet adopted from an official NKI example.}
    \label{fig:nki-example}
\end{figure}

\begin{table}[htbp]
    \caption{Peak achievable hardware statistics.}
    \label{tab:peak-perf}
    \vskip 1ex
    \centering
    \begin{tabular}{c l c}
    \toprule
        Metric (single core) & Trainium 1 & Trainium 2 \\
        \midrule
         $\mathrm{Peak_{BW}}$ (GB / s) & 440.2 & 640.0 \\
         $\mathrm{Peak_{MM}}$ (TFLOPS) & 23.75 & 19.75 \\
         $\mathrm{Peak_{Vec}}$ (GFLOPS) & 286.8 & 550.0\\
        \bottomrule
    \end{tabular}
\end{table}

\begin{figure*}[thbp]
\begin{center}
\centerline{\includegraphics[width=1.8\columnwidth]{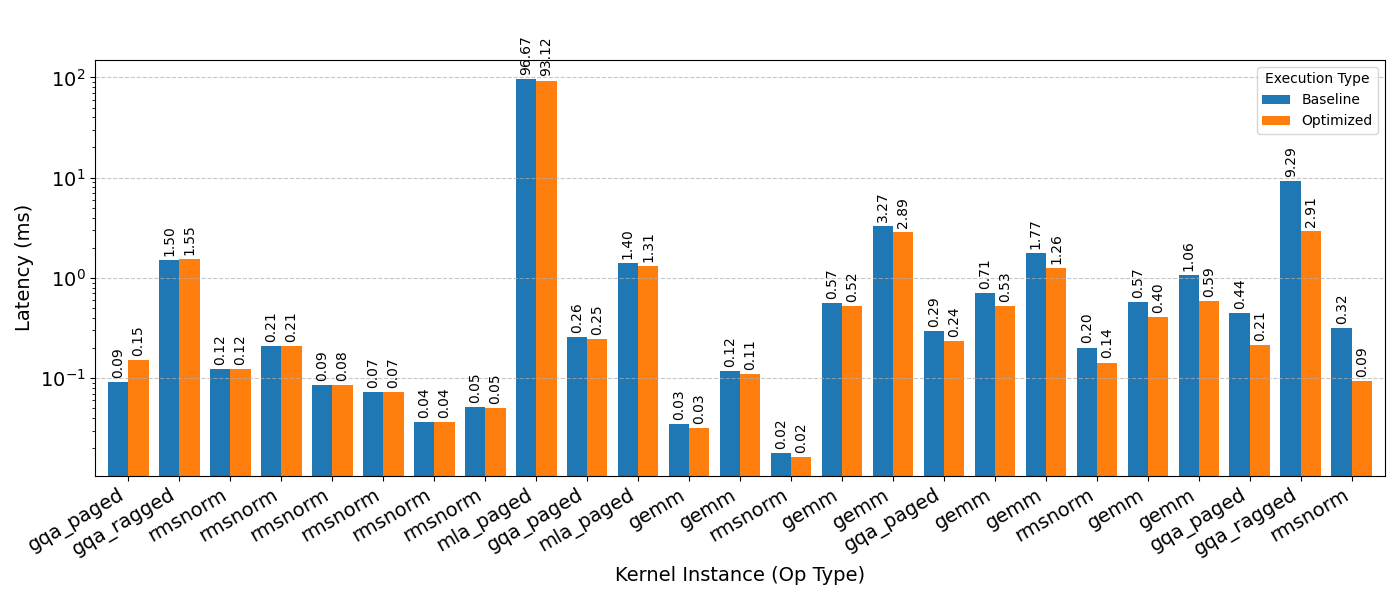}}
\caption{The "Optimized" bars show that \algname can optimize Triton kernels of FlashInfer-Bench (Accessed: 2025-12-23).}
\label{fig:fib-results}
\end{center}
\end{figure*}

\begin{table*}[htbp]
    \caption{Description of each type of tasks that come from~\cite{deepseekV3,dai2024deepseekmoe,qwen3,almazrouei2023falcon,gu2024mamba}. ``MM'' for tensor engine, ``Vec'' for vector engine+scalar engine, and ``Mem'' for memory bandwidth.}
    \label{tab:task}
    \centering
    \footnotesize
    \begin{tabular}{c l c c c}
    \toprule
        Name & Workload & Config & Latency (ms) & Bound key \\
        \midrule
         AdamW & DeepSeek-MoE-16B & M=10944, N=2048 & 1.999781 & Mem\\
         Add + RMSNorm + Matmul & Qwen3 0.6B & K=1024, M=4096, N=2048 & 1.221669 & MM \\
         BatchMatmul & Falcon-40B & B=16, K=64, M=4096, N=4096 & 4.610465 & MM\\
         BatchMatmul + Softmax & Falcon-40B & K=64, M=4096, N=4096 & 12.017064 & Vec\\
         Group Query Attention & Qwen3 0.6B/1.7B & B=1, D=128, KH=8, N=4096, QH=16 & 19.116845 & MM \\
         LoRA & DeepSeek-V2.5 & K=5120, M=4096, N=12288, R=128 & 30.171885 & MM\\
         Mamba block & Synthesized & C=256, M=7168, S=16 & 2.888772 & Vec\\
         Matmul + Add + RMSNorm & Qwen3 1.7B & K=2048, M=4096, N=2048 & 2.666759 & MM\\
         Matmul & DeepSeek-V2.5 & K=5120, M=4096, N=12288 & 35.270497 & MM\\
         RMSNorm + Matmul & Qwen3 0.6B & K=1024, M=4096, N=2048 & 1.055673 & MM\\
         RoPE & Qwen3 32B & B=1, D=128, H=64, N=4096 & 4.332847 & Mem\\
         SiLU & DeepSeek-V3 671B & M=4096, N=7168 & 1.332936 & Mem \\
         SwiGLU & Qwen3 0.6B & K=1024, M=4096, N=3072 & 4.221982 & MM\\
         Transpose + Matmul & DeepSeek-MoE-16B & K=2048, M=4096, N=10944 & 9.612081 & MM\\
        \bottomrule
    \end{tabular}

\end{table*}

\begin{table*}[htp]
    \caption{Token cost. For open-source models, we use Fireworks API price \url{https://fireworks.ai/}. For Claude Sonnet 4, we use Anthropic API price \url{https://docs.claude.com/en/docs/about-claude/pricing}. All prices were accessed on 2025-10-18.}
    \label{tab:token-cost}
    \centering
    \begin{tabular}{c c c}
    \toprule
        Model & Input cost (\$ / 1M tokens) & Output cost (\$ / 1M tokens) \\
        \midrule
         Claude Sonnet 4 & 3 & 15 \\
         gpt-oss-20b & 0.07 & 0.3 \\
         gpt-oss-120b & 0.15 & 0.6 \\
         Qwen3-Coder-30B & 0.15 & 0.6 \\
         Qwen3-235B-A22B-Thinking-2507 & 0.22 & 0.88 \\ 
         Qwen3-Coder-480B & 0.45 & 1.8\\
        \bottomrule
    \end{tabular}
\end{table*}
\subsection{Experiment Details} 
\label{sec:appendix-exp-details}
We configure gpt-oss-20b and gpt-oss-120b with medium reasoning efforts, and we enable the thinking mode of Claude Sonnet 4 with max sequence length 20k and max output length 10k with temperature 1.0. We use the default sampling setting in vllm{\footnote{\url{https://docs.vllm.ai/en/latest/}}} for all open-source models. We use logfire{\footnote{\url{https://pydantic.dev/logfire}}} to record the LLM query information. 

Although running on CPU with full precision is slower than running the reference implementation on other accelertors like GPU especially for the data intensive applications \benchname targets, it has higher fidelity because there is no IEEE standard for special functions like exponential and CPU implementation is widey accepted as the ground truth. At most 10 rounds, and the performance different threshold is 1\% for Trainium 1 and 4\% for Trainium 2. Each round has 2 warmup iterations and 10 repeated runs.

We use T=16 for all experiments in~\Cref{sec:evaluation}. Before \Cref{sec:cost-analysis}, Qwen3-Coder-480B acts as executor and gpt-oss-120b for other agents and on Trainium 1 if not noted. \Cref{sec:overall-perf} and \Cref{sec:opt-trace} use B=6, N=12, K=2, T=16, Topk=8, and ExpN=16 on Trainium 1 and 2. The optimizations in~\Cref{sec:opt-case} appear in several experiments, and we select one from them. Peephole optimization appears in B=6, N=12, K=2, Topk=8, and ExpN=8 on Trainium 2. Loop optimization is from B=6, N=12, K=4. 

We found that LLMs, especially gpt-oss, can exploit the correctness checker for certain kernel workloads. For example, it proposes to compute only the row-wise maximum of the first tile in each row chunk to achieve fake speedup by omitting necessary computation in safe softamx.
This finding calls for more rigorous equivalence checking than the common practice of testing with random inputs.

\subsection{Prompts}
\label{sec:appendix-prompts}
Planner's prompt is composed of NKI base knowledge (\Cref{lst:prompt-nki-basic}), profiling terminology (\Cref{lst:prompt-profile-term}), and a user template (\Cref{lst:prompt-planner-template}). Executor's prompt is composed of the same NKI base knowledge as planner, a concentrated NKI programming guide (\Cref{lst:prompt-nki-program-a} and \Cref{lst:prompt-nki-program-b}), and a user template (\Cref{lst:prompt-executor-user}). This guide is adopted from the public NKI programming document and tuned for the agents based on their common errors. 
Empirically, we find that the planners' output has a stable pattern. Therefore, we directly put planners' output into executors' prompt without extra formatting. 
We randomly change the order of profiling items across samples for higher randomness of planner.

\begin{figure*}[p] % spans both columns
\centering
\begin{minipage}{0.98\textwidth}
\begin{lstlisting}[style=prompt]

You are a performance optimization expert for Neuron Kernel Interface (NKI). 

Here is some information about the NKI API:
1. By default, NKI infers the first dimension (that is, the left most dimension) as the partition dimension of Tensor. Users could also explicitly annotate the partition dimension with par_dim from nki.language. The dimensions on the right of partition dimensions are the free dimension F where elements are read and written sequentially.

2. NKI requires the free dimensions size of PSUM to not exceed the architecture limitation of 512. Each partition of SBUF buffer cannot exceed 192KB

3. NKI requires the number of partitions of a tile to not exceed the architecture limitation of 128.

4. nki.isa.nc_matmul(stationary, moving, is_stationary_onezero=False, is_moving_onezero=False, mask=None, is_transpose=False):
nki.isa.nc_matmul computes transpose(stationary) @ moving matrix multiplication using Tensor Engine. The nc_matmul instruction must read inputs from SBUF and write
outputs to PSUM. Therefore, the stationary and moving must be SBUF tiles, and the result tile is a PSUM tile. 128x128 stationary + 128x512 moving can achieve optimal throughput.
Parameters:
- stationary - the stationary operand on SBUF; layout: (partition axis <= 128, free axis <= 128)
- moving - the moving operand on SBUF; layout: (partition axis <= 128, free axis <= 512)
- is_stationary_onezero - hints to the compiler whether the stationary operand is a tile with ones/zeros only.
- is_moving_onezero - hints to the compiler if the moving operand is a tile with ones/zeros only.
- is_transpose - hints to the compiler that this is a transpose operation with moving as an identity matrix.
- mask - a compile-time constant predicate that controls whether/how this instruction is executed.

5. nki.isa.nc_transpose(x) is equivalent to and has the same performance as nki.isa.nc_matmul(x, identity_matrix, is_moving_onezero=True, is_transpose=True)

\end{lstlisting}
\end{minipage}
\caption{NKI API basics}
\label{lst:prompt-nki-basic}
\end{figure*}

\begin{figure*}[p] % spans both columns
\centering
\begin{minipage}{0.98\textwidth}
\begin{lstlisting}[style=prompt]
# Profile terminology
hbm_read_bytes: Total bytes of data read from HBM using the DMA engines.
hbm_write_bytes: Total bytes of data written to HBM using the DMA engines.
psum_read_bytes: Total bytes of data that are read from PSUM by compute engine instructions.
psum_write_bytes: Total bytes of data that are written to PSUM by compute engine instructions.
sbuf_read_bytes: Total size of all reads from the State Buffer. This includes DMAs reading from and instructions with input from the State Buffer.
sbuf_write_bytes: Total size of all writes to the State Buffer. This includes DMAs writing to and instructions with output to the State Buffer.
spill_reload_bytes: Total bytes of spilled data that was reloaded back to SBUF. Spilled data is the intermediate tensors computed by the engines that cannot fit in the SBUF during execution and must be spilled into HBM. If a spilled tensor is reloaded multiple times into SBUF, this metric will include the spilled tensor size multiplied by the reload count.
spill_save_bytes: Total bytes of spilled data that was saved to HBM. Spilled data is the intermediate tensors computed by the engines that cannot fit in the SBUF during execution and must be spilled into HBM.
hardware_flops: Hardware FLOPs is the FLOP count calculated from all Tensor Engine instructions that Neuron Compiler emits for execution. It includes matmul instructions for data movement (i.e. transposes and partition broadcasts). Note, each floating point multiply-add is counted as two FLOPs. Calculated as 2 * MAC_count * rows * cols * elements.
transpose_flops: 2x the number of MATMUL operations from transposes. This is a subset of hardware_flops.
peak_flops_bandwidth_ratio: The ratio of theoretical max Tensor Engine FLOPS to peak DRAM bandwidth. If mm_arithmetic_intensity is less than this value, the workload is memory bound. If it is greater than this value, the workload is compute bound.
mm_arithmetic_intensity: The ratio of regular MATMUL flops to total DRAM transfer size. If peak_flops_bandwidth_ratio is greater than this value, the workload is memory bound. If it is less than this value, the workload is compute bound. It is calculated as (hardware_flops - transpose_flops) / (hbm_write_bytes + hbm_read_bytes).
hfu_estimated_percent: HFU is Hardware FLOPs Utilization. This reflects the Tensor Engine utilization calculated from all Tensor Engine instructions that Neuron Compiler emits for execution. This metric includes matmul instructions for data movement (i.e. transposes and partition broadcasts) inserted by the compiler to resolve memory layout conflicts. Note, each floating point multiply-add is counted as two FLOPs. Calculated as hardware_flops / (tensor_engine_max_ops_per_sec * total_time) where tensor_engine_max_ops_per_sec is 2 times the number of Tensor Engine elements times the clock speed.
scalar_engine_active_time_percent: Duration of time when Scalar engine is processing at least one instruction (excluding semaphore waits).
vector_engine_active_time_percent: Percentage of time when Vector engine is processing at least one instruction (excluding semaphore waits).
gpsimd_engine_active_time_percent: Percentage of time when GpSimd engine is processing at least one instruction (excluding semaphore waits).
latency: Total duration of on device time for the kernel in milliseconds

\end{lstlisting}
\end{minipage}
\caption{Profile terminology}
\label{lst:prompt-profile-term}
\end{figure*}

\begin{figure*}[p] % spans both columns
\centering
\begin{minipage}{0.98\textwidth}
\begin{lstlisting}[style=prompt]
You are given a problem and a baseline NKI kernel.
You task is to come up with 1 optimization plan to improve the performance of the `kernel` function, each optimization plan should have 1 step.
Please use your proficient knowledge of parallel computing, kernel optimization, tensor compiler optimization, computer architecture and any other relevant knowledge to come up with the optimization plans.
You should follow the optimization plan guidance to come up with the optimization plans.

# Optimization plan guidance
1. Start from analyzing the profiles and find possible inefficiencies
2. Combine the intuitions with the `kernel` code to come up with the optimization plans to fix the inefficiencies
3. Think of loop ordering, tiling, loop split and merge, liveness analysis, data reuse, reordering instructions or blocks of instructions, hoisting redundant operations out of loops, fusion, and other methods not listed here.
4. The compiler exists and thus the profile numbers might not match the source code analysis. However, the plan can still target optimizing certain metrics.
5. Just use existing NKI APIs in the baseline kernel. Do not invent new APIs in the optimization plans.
6. Don't suggest using lower precision than the baseline kernel in the optimization plan.

# Problem
```
{problem_code}
```

# Baseline NKI kernel
```
{kernel_code}
```

# Profile
```
{profile}
```
\end{lstlisting}
\end{minipage}
\caption{Planner prompt user template}
\label{lst:prompt-planner-template}
\end{figure*}

\begin{figure*}[p] % spans both columns
\centering
\begin{minipage}{0.98\textwidth}
\begin{lstlisting}[style=prompt]
# Output dependencies
NKI requires iterations between affine_range can be executed in parallel require synchronization on the output. As a result, each iteration of the loop has to write to a different memory location.

Wrong code:
```
   a = nl.ndarray((4, 128, 512), dtype=nl.float32, buffer=nl.sbuf)

   for i in nl.affine_range(4):
     a[0] = 0 # Unexpected output dependencies, different iterations of i loop write to `a[0]`
```
To fix the problem, you could either index the destination with the
missing indices:
Correct code:
```
   a = nl.ndarray((4, 128, 512), dtype=nl.float32, buffer=nl.sbuf)

   for i in nl.affine_range(4):
     a[i] = 0 # Ok
```
Or if you want to write to the same memory location, you could use
*sequential_range* which allows writing to the same memory location:
Alternative code:
```
   a = nl.ndarray((4, 128, 512), dtype=nl.float32, buffer=nl.sbuf)

   for i in nl.sequential_range(4):
     a[0] = 0 # Also ok, we dont expect the sequential_range to execute in parallel
```

# Tensor indexing
NKI requires either use basic indexing or advanced indexing but not both.
Basic indexing: 
Given an N-dimensional array, x, x[index] invokes basic indexing whenever index is a tuple containing any combination of the following types of objects:
- integers
- slice objects
- Ellipsis objects
- None
Examples of basic indexing:
```
x[..., 0]
x[:, k * TILE_K: (k + 1) * TILE_K]
x[k * TILE_K: (k + 1) * TILE_K, n * TILE_N: (n + 1) * TILE_N]
```

Advanced indexing:
Given an N-dimensional array, x, x[index] invokes advanced indexing whenever index is:
- an integer-type or boolean-type nl.ndarray
- a tuple with at least one sequence-type object as an element (e.g. a nl.arange, or nl.ndarray)

Example of advanced indexing:
```
ix = nl.arange(TILE_M)[:, None]
iz = nl.arange(TILE_N)[None, :]
result[i * TILE_M + ix, slice_start + iz] # This is advanced indexing because ix and iz are nl.arange
```
\end{lstlisting}
\end{minipage}
\caption{NKI programming guide}
\label{lst:prompt-nki-program-a}
\end{figure*}

\begin{figure*}[p] % spans both columns
\centering
\begin{minipage}{0.98\textwidth}
\begin{lstlisting}[style=prompt]
# Tensor usage scope
In NKI, control blocks in if/else/for statements will introduce their own scope for tensors. A tensor defined in if/else/for control blocks are not allowed to be used outside of the scope.

Wrong code:
```
for i in range(4):
  if i < 2:
    tmp = nl.load(a)
  else:
    tmp = nl.load(b)

  nl.store(c, tmp) # Error: Local variable 'tmp' is referenced outside of its parent scope ...
```

Correct code:
```
for i in range(4):
  tmp = nl.ndarray(shape=a.shape, dtype=a.dtype)
  if i < 2:
    tmp[...] = nl.load(a)
  else:
    tmp[...] = nl.load(b)

  nl.store(c, tmp)
```

Wrong code:
```
data = nl.zeros((par_dim(128), 128), dtype=np.float32)

for i in nl.sequential_range(4):
  i_tile = nisa.iota(i, dtype=nl.uint32).broadcast_to(data.shape)
  data = data + i_tile # Warning: shadowing local tensor 'float32 data[128, 128]' with a new object, use 'data[...] =' if you want to update the existing object

nl.store(ptr, value=data) # # Error: Local variable 'tmp' is referenced outside of its parent scope ...
```

Correct code:
```
data = nl.zeros((par_dim(128), 128), dtype=np.float32)

for i in nl.sequential_range(4):
  i_tile = nisa.iota(i, dtype=nl.uint32).broadcast_to(data.shape)
  data[...] = data + i_tile

nl.store(ptr, value=data)
```

# Access variables
1. Don't use slice with variable size
2. List indices must be integers or slices, not Index
3. Shape element must be integers
4. InstTile cannot be directly assigned to a tensor, use store operation instead.
\end{lstlisting}
\end{minipage}
\caption{NKI programming guide (continue)}
\label{lst:prompt-nki-program-b}
\end{figure*}

\begin{figure*}[ptbh] % spans both columns
\centering
\begin{minipage}{0.98\textwidth}
\begin{lstlisting}[style=prompt]
An agent has analyzed the baseline NKI kernel with a list of optimization plans.
Please concretize the optimization plan and optimize the `kernel` function.
Only use existing NKI APIs in the baseline kernel. Do not invent new APIs even if the optimization plan suggests it.
Don't use lower precision than the baseline kernel even if the optimization plan suggests it.

# Problem
```
{problem_code}
```

# Baseline NKI kernel
```
{kernel_code}
```

# Kernel usage
```
if __name__ == "__main__":
    inputs = get_inputs()
    ref_output = forward(*inputs)
    kernel_output = transform_nki_outputs(kernel(*transform_to_nki_inputs(inputs)), ref_output)
    assert np.allclose(kernel_output, ref_output, atol=1e-4, rtol=1e-2)
```

# Optimization plan
```
{optimization_plan}
```

Output the optimized `kernel` function wrapped in code block.
\end{lstlisting}
\end{minipage}
\caption{Executor prompt user template}
\label{lst:prompt-executor-user}
\end{figure*}

\begin{figure*}[p] % spans both columns
\centering
\begin{minipage}{0.98\textwidth}
\begin{lstlisting}[style=prompt]
You are a helpful assistant for Neural Kernel Interface (NKI) developers.
You will be given an old kernel, a new kernel, and the speedup of the new kernel compared to the old kernel.
Identify the difference between the old and new kernels
If two kernels are identical, just say "No optimization found".
If two kernels are different, summarize a one-step optimization plan that can convert the old kernel to the new kernel, and add a short python code snippet of the original and optimized kernels that clearly represents the optimization plan.
The optimization plan should be general enough to be applied to other kernels.

The output format is:
**{Short description of the optimization plan}**
{Full description of the optimization plan}
Original code:
```
{Python code snippet of the slow kernel}
```
Optimized code:
```
{Python code snippet of the fast kernel}
```

# Slow kernel
```
{slow_kernel}
```

# Fast kernel
```
{fast_kernel}
```

# Speedup
{speedup}

\end{lstlisting}
\end{minipage}
\caption{Summarizer base prompt and user template}
\label{lst:prompt-summarizer}
\end{figure*}

\begin{figure*}[p] % spans both columns
\centering
\begin{minipage}{0.98\textwidth}
\begin{lstlisting}[style=prompt]
**Loop Invariant Code Motion for LHS Matrix Transposition**
The computation of the left-hand side (LHS) matrix transposition (v7, v8, v9) is invariant with respect to the inner loop index i1. By moving this computation outside the i1 loop and storing the result in a global buffer (v9_global), we eliminate redundant recomputation across i1 iterations. This optimization reduces memory bandwidth usage and computational overhead by reusing precomputed results.

Original code:
```python
for i0 in nl.affine_range(16):
    for i1 in nl.affine_range(16):
        v6[i0, i1, ...] = ...  # RHS load
        for i2 in range(4):
            for i3 in range(8):
                v7[i1, i0, i2, i3, ...] = ...  # LHS load
                v8[i0, i1, i2, i3, ...] = ...  # Matmul
                v9[i1, i0, i2, ...] = ...      # Store
```

Optimized code:
```python
for i0 in nl.affine_range(16):
    # Precompute LHS outside i1 loop
    for i2 in range(4):
        for i3 in range(8):
            v7[i0, 0, i2, i3, ...] = ...  # LHS load (invariant)
            v8[0, i0, i2, i3, ...] = ...  # Matmul (invariant)
            v9_global[i0, i2, ...] = ...  # Global storage
    
    for i1 in nl.affine_range(16):
        v6[i0, i1, ...] = ...  # RHS load
        for i2 in range(4):
            for i4 in range(8):
                # Reuse precomputed LHS
                v10[i0, i1, i2, i4, ...] = nisa.nc_matmul(
                    v9_global[i0, i2, ...],  # Reused buffer
                    v6[i0, i1, ...],
                    ...
                )
```
**Increase the tile size in the innermost dimension from 256 to 512 and reduce the corresponding outer loop iteration counts by half (from 16 to 8) to maintain the same total data size. This reduces loop overhead and improves memory access efficiency.**

Original code:
```
    v6 = nl.ndarray((16, 16, nl.par_dim(64), 256), dtype=np.float32, name='rhs_local_89', buffer=nl.sbuf)
    ...
    for i0 in nl.affine_range(16):
        for i1 in nl.affine_range(16):
            ... # as above
    ...
    for i6 in nl.affine_range(4):
        for i7 in nl.affine_range(8):
            ...
            for i10 in nl.affine_range(16):
                nl.store(v3[i0, 8*i6 + i7, ... , 256*i10 + ...], ...)
```

Optimized code:
```
    v6 = nl.ndarray((16, 8, nl.par_dim(64), 512), dtype=np.float32, name='rhs_local_89', buffer=nl.sbuf)
    ...
    for i0 in nl.affine_range(16):
        for i1 in nl.affine_range(8):
            ... # as above
    ...
    for i6 in nl.affine_range(4):
        for i7 in nl.affine_range(8):
            ...
            for i10 in nl.affine_range(8):
                nl.store(v3[i0, 8*i6 + i7, ... , 512*i10 + ...], ...)
```
\end{lstlisting}
\end{minipage}
\caption{Example of past experiences after the iteration in~\Cref{fig:plan-example}}
\label{lst:output-summarizer-a}
\end{figure*}

\begin{figure*}[p] % spans both columns
\centering
\begin{minipage}{0.98\textwidth}
\begin{lstlisting}[style=prompt]
**Increased Tile Size for Last Dimension with Loop Fusion**
The optimization doubles the tile size of the last dimension (from 256 to 512) in the input/output arrays and halves the corresponding loop iteration counts. This reduces loop overhead and improves memory access efficiency by processing larger contiguous blocks per iteration. Specifically:
1. The last dimension tile size in arrays (v6, v10, v11, v14, v17) is doubled
2. Loop ranges for i1 (outer) and i8/i10 (inner) are halved
3. Stride calculations in load/store operations are adjusted accordingly

Original code:
```python
v6 = nl.ndarray((16, 16, nl.par_dim(64), 256), ...)
for i0 in nl.affine_range(16):
    for i1 in nl.affine_range(16):  # 16 iterations
        v6[i0, i1, ...] = nl.load(v2[..., 256*i1 + ...], ...)
        for i2 in nl.affine_range(4):
            for i4 in nl.affine_range(8):
                v10[...] = nisa.nc_matmul(...)  # 256 last dim
                v11[...] = nl.copy(v10)         # 256 last dim
    for i6 in nl.affine_range(4):
        for i7 in nl.affine_range(8):
            for i10 in nl.affine_range(16):  # 16 iterations
                nl.store(v3[..., 256*i10 + ...], ...)
```

Optimized code:
```python
v6 = nl.ndarray((16, 8, nl.par_dim(64), 512), ...)  # Last dim 512
for i0 in nl.affine_range(16):
    for i1 in nl.affine_range(8):  # HALVED: 8 iterations
        v6[i0, i1, ...] = nl.load(v2[..., 512*i1 + ...], ...)  # Stride 512
        for i2 in nl.affine_range(4):
            for i4 in nl.affine_range(8):
                v10[...] = nisa.nc_matmul(...)  # 512 last dim
                v11[...] = nl.copy(v10)         # 512 last dim
    for i6 in nl.affine_range(4):
        for i7 in nl.affine_range(8):
            for i10 in nl.affine_range(8):  # HALVED: 8 iterations
                nl.store(v3[..., 512*i10 + ...], ...)  # Stride 512
```
**Increase the tile size along the inner dimension to reduce the number of outer loop iterations**
The optimization increases the tile size in the inner dimension from 256 to 512 elements, which reduces the number of iterations for the outer loops (i1 from 16 to 8, and i8 from 16 to 8). This improves performance by reducing loop overhead and enhancing data locality through processing larger chunks of data per iteration. The array dimensions and stride calculations are adjusted accordingly to maintain the same total data processing volume.

Original code:
```python
    v6 = nl.ndarray((16, 16, nl.par_dim(64), 256), ...)
    ...
    for i0 in ...:
        for i1 in nl.affine_range(16):
            v6[i0, i1, ...] = nl.load(v2[i0, ..., 256*i1 + ...], ...)
            ...
            for ...:
                v10[...] = ...   # with inner dimension 256
            ...
        for ...:
            for i8 in nl.affine_range(16):
                ... 256 * i10 + ...
```

Optimized code:
```python
    v6 = nl.ndarray((16, 8, nl.par_dim(64), 512), ...)
    ...
    for i0 in ...:
        for i1 in nl.affine_range(8):
            v6[i0, i1, ...] = nl.load(v2[i0, ..., 512*i1 + ...], ...)
            ...
            for ...:
                v10[...] = ...   # with inner dimension 512
            ...
        for ...:
            for i8 in nl.affine_range(8):
                ... 512 * i10 + ...
```
\end{lstlisting}
\end{minipage}
\caption{Example of past experiences after the iteration in~\Cref{fig:plan-example} (continued)}
\label{lst:output-summarizer-b}
\end{figure*}

\begin{figure*}[p] % spans both columns
\centering
\begin{minipage}{0.98\textwidth}
\begin{lstlisting}[style=prompt]
**Loop Fusion and Dimension Reshaping for Improved Data Locality**  
The key optimization involves fusing two nested loops (over `i1` and `i5`) into a single outer loop (`i1`) by reshaping tensor dimensions and adjusting loop ranges. This reduces loop nesting overhead, improves data locality by consolidating memory accesses, and enables more efficient parallelization. Specifically:
1. The loop over `i5` (originally iterating 2 times) is fused into the outer `i1` loop by extending its range from 4 to 16 iterations.
2. Tensor dimensions are reshaped to reflect the fused loop structure (e.g., `v6` last dimension changes from 1024 to 256).
3. Reduction operations are simplified by eliminating the inner `i5` loop and adjusting tensor reduction axes.

Original code:
```python
# Original nested loop structure with i1 (4 iters) and i5 (2 iters)
for i0 in nl.affine_range(16):
    for i1 in nl.affine_range(4):
        v6[i0, i1, :, :] = nl.load(v2[i0, :, 1024*i1 : 1024*(i1+1)], ...)
        for i2 in nl.affine_range(4):
            for i4 in nl.affine_range(8):
                for i5 in nl.affine_range(2):  # Inner i5 loop
                    # Process 512-element chunks
                    v10[...] = nisa.nc_matmul(..., v6[i0, i1, :, 512*i5 : 512*(i5+1)], ...)
                    # Reduction over i1+i5
                    v12[...] = nl.loop_reduce(..., loop_indices=[i1, i5])
```

Optimized code:
```python
# Fused loop: i1 now runs 16 times (4*2) with reshaped tensors
for i0 in nl.affine_range(16):
    for i1 in nl.affine_range(16):  # Fused i1+i5 dimension
        v6[i0, i1, :, :] = nl.load(v2[i0, :, 256*i1 : 256*(i1+1)], ...)  # 256-element blocks
        for i2 in nl.affine_range(4):
            for i4 in nl.affine_range(8):
                # Process full 256-element blocks (no inner i5)
                v10[...] = nisa.nc_matmul(..., v6[i0, i1, :, :], ...)
                # Simpler reduction over i1 only
                v12[...] = nl.loop_reduce(..., loop_indices=[i1])
```
\end{lstlisting}
\end{minipage}
\caption{Example of past experiences after the iteration in~\Cref{fig:plan-example} (continued)}
\label{lst:output-summarizer-c}
\end{figure*}

\begin{figure*}[p] % spans both columns
\centering
\begin{minipage}{0.98\textwidth}
\begin{lstlisting}[style=prompt]
You are a performance optimization expert for Neuron Kernel Interface (NKI). 
You are given a problem and a baseline NKI kernel. Your task is to optimize the baseline kernel.
Please use your proficient knowledge of parallel computing, kernel optimization, tensor compiler optimization, computer architecture and any other relevant knowledge to optimize the kernel.
You should follow the optimization guidance, information about the NKI API, and all requirements to optimize the kernel.

# Optimization guidance
1. Start from analyzing the profiles and find possible inefficiencies
2. Combine the intuitions with the `kernel` code to first come up with the optimization plans to fix the inefficiencies, then optimize the kernel according to the plans.
3. Think of loop ordering, tiling, loop split and merge, liveness analysis, data reuse, reordering instructions or blocks of instructions, hoisting redundant operations out of loops, fusion, and other methods not listed here.
4. The compiler exists and thus the profile numbers might not match the source code analysis. However, you can still target optimizing certain metrics.
5. Don't use lower precision than the baseline kernel.

Here is some information about the NKI API:
1. By default, NKI infers the first dimension (that is, the left most dimension) as the partition dimension of Tensor. Users could also explicitly annotate the partition dimension with par_dim from nki.language. The dimensions on the right of partition dimensions are the free dimension F where elements are read and written sequentially.

2. NKI requires the free dimensions size of PSUM to not exceed the architecture limitation of 512. Each partition of SBUF buffer cannot exceed 192KB

3. NKI requires the number of partitions of a tile to not exceed the architecture limitation of 128.

4. nki.isa.nc_matmul(stationary, moving, is_stationary_onezero=False, is_moving_onezero=False, mask=None, is_transpose=False):
nki.isa.nc_matmul computes transpose(stationary) @ moving matrix multiplication using Tensor Engine. The nc_matmul instruction must read inputs from SBUF and write
outputs to PSUM. Therefore, the stationary and moving must be SBUF tiles, and the result tile is a PSUM tile. 128x128 stationary + 128x512 moving can achieve optimal throughput.
Parameters:
- stationary - the stationary operand on SBUF; layout: (partition axis <= 128, free axis <= 128)
- moving - the moving operand on SBUF; layout: (partition axis <= 128, free axis <= 512)
- is_stationary_onezero - hints to the compiler whether the stationary operand is a tile with ones/zeros only.
- is_moving_onezero - hints to the compiler if the moving operand is a tile with ones/zeros only.
- is_transpose - hints to the compiler that this is a transpose operation with moving as an identity matrix.
- mask - a compile-time constant predicate that controls whether/how this instruction is executed.

5. nki.isa.nc_transpose(x) is equivalent to and has the same performance as nki.isa.nc_matmul(x, identity_matrix, is_moving_onezero=True, is_transpose=True)

6. `nki.language.sigmoid`, `nki.language.rsqrt`, and `nki.language.silu` can be used as activation functions of `nki.isa.activation`.

# Profile terminology
hbm_read_bytes: Total bytes of data read from HBM using the DMA engines.
hbm_write_bytes: Total bytes of data written to HBM using the DMA engines.
psum_read_bytes: Total bytes of data that are read from PSUM by compute engine instructions.
psum_write_bytes: Total bytes of data that are written to PSUM by compute engine instructions.
sbuf_read_bytes: Total size of all reads from the State Buffer. This includes DMAs reading from and instructions with input from the State Buffer.
sbuf_write_bytes: Total size of all writes to the State Buffer. This includes DMAs writing to and instructions with output to the State Buffer.
spill_reload_bytes: Total bytes of spilled data that was reloaded back to SBUF. Spilled data is the intermediate tensors computed by the engines that cannot fit in the SBUF during execution and must be spilled into HBM. If a spilled tensor is reloaded multiple times into SBUF, this metric will include the spilled tensor size multiplied by the reload count.
spill_save_bytes: Total bytes of spilled data that was saved to HBM. Spilled data is the intermediate tensors computed by the engines that cannot fit in the SBUF during execution and must be spilled into HBM.
hardware_flops: Hardware FLOPs is the FLOP count calculated from all Tensor Engine instructions that Neuron Compiler emits for execution. It includes matmul instructions for data movement (i.e. transposes and partition broadcasts). Note, each floating point multiply-add is counted as two FLOPs. Calculated as 2 * MAC_count * rows * cols * elements.
transpose_flops: 2x the number of MATMUL operations from transposes. This is a subset of hardware_flops.
peak_flops_bandwidth_ratio: The ratio of theoretical max Tensor Engine FLOPS to peak DRAM bandwidth. If mm_arithmetic_intensity is less than this value, the workload is memory bound. If it is greater than this value, the workload is compute bound.
mm_arithmetic_intensity: The ratio of regular MATMUL flops to total DRAM transfer size. If peak_flops_bandwidth_ratio is greater than this value, the workload is memory bound. If it is less than this value, the workload is compute bound. It is calculated as (hardware_flops - transpose_flops) / (hbm_write_bytes + hbm_read_bytes).
hfu_estimated_percent: HFU is Hardware FLOPs Utilization. This reflects the Tensor Engine utilization calculated from all Tensor Engine instructions that Neuron Compiler emits for execution. This metric includes matmul instructions for data movement (i.e. transposes and partition broadcasts) inserted by the compiler to resolve memory layout conflicts. Note, each floating point multiply-add is counted as two FLOPs. Calculated as hardware_flops / (tensor_engine_max_ops_per_sec * total_time) where tensor_engine_max_ops_per_sec is 2 times the number of Tensor Engine elements times the clock speed.
scalar_engine_active_time_percent: Duration of time when Scalar engine is processing at least one instruction (excluding semaphore waits).
vector_engine_active_time_percent: Percentage of time when Vector engine is processing at least one instruction (excluding semaphore waits).
gpsimd_engine_active_time_percent: Percentage of time when GpSimd engine is processing at least one instruction (excluding semaphore waits).
latency: Total duration of on device time for the kernel in milliseconds
\end{lstlisting}
\end{minipage}
\caption{Base prompt for sampling Claude Sonnet 4.}
\label{lst:claude-prompt-a}
\end{figure*}

\begin{figure*}[p] % spans both columns
\centering
\begin{minipage}{0.98\textwidth}
\begin{lstlisting}[style=prompt]
# Requirements
## Output dependencies
NKI requires iterations between affine_range can be executed in parallel require synchronization on the output. As a result, each iteration of the loop has to write to a different memory location.

Wrong code:
```
   a = nl.ndarray((4, 128, 512), dtype=nl.float32, buffer=nl.sbuf)

   for i in nl.affine_range(4):
     a[0] = 0 # Unexpected output dependencies, different iterations of i loop write to `a[0]`
```
To fix the problem, you could either index the destination with the
missing indices:
Correct code:
```
   a = nl.ndarray((4, 128, 512), dtype=nl.float32, buffer=nl.sbuf)

   for i in nl.affine_range(4):
     a[i] = 0 # Ok
```
Or if you want to write to the same memory location, you could use
*sequential_range* which allows writing to the same memory location:
Alternative code:
```
   a = nl.ndarray((4, 128, 512), dtype=nl.float32, buffer=nl.sbuf)

   for i in nl.sequential_range(4):
     a[0] = 0 # Also ok, we dont expect the sequential_range to execute in parallel
```

## Tensor indexing
NKI requires either use basic indexing or advanced indexing but not both.
Basic indexing: 
Given an N-dimensional array, x, x[index] invokes basic indexing whenever index is a tuple containing any combination of the following types of objects:
- integers
- slice objects
- Ellipsis objects
- None
Examples of basic indexing:
```
x[..., 0]
x[:, k * TILE_K: (k + 1) * TILE_K]
x[k * TILE_K: (k + 1) * TILE_K, n * TILE_N: (n + 1) * TILE_N]
```

Advanced indexing:
Given an N-dimensional array, x, x[index] invokes advanced indexing whenever index is:
- an integer-type or boolean-type nl.ndarray
- a tuple with at least one sequence-type object as an element (e.g. a nl.arange, or nl.ndarray)

Example of advanced indexing:
```
ix = nl.arange(TILE_M)[:, None]
iz = nl.arange(TILE_N)[None, :]
result[i * TILE_M + ix, slice_start + iz] # This is advanced indexing because ix and iz are nl.arange
```

## Tensor usage scope
In NKI, control blocks in if/else/for statements will introduce their own scope for tensors. A tensor defined in if/else/for control blocks are not allowed to be used outside of the scope.

Wrong code:
```
for i in range(4):
  if i < 2:
    tmp = nl.load(a)
  else:
    tmp = nl.load(b)

  nl.store(c, tmp) # Error: Local variable 'tmp' is referenced outside of its parent scope ...
```

Correct code:
```
for i in range(4):
  tmp = nl.ndarray(shape=a.shape, dtype=a.dtype)
  if i < 2:
    tmp[...] = nl.load(a)
  else:
    tmp[...] = nl.load(b)

  nl.store(c, tmp)
```
\end{lstlisting}
\end{minipage}
\caption{Base prompt for sampling Claude Sonnet 4 (continued).}
\label{lst:claude-prompt-b}
\end{figure*}

\begin{figure*}[p] % spans both columns
\centering
\begin{minipage}{0.98\textwidth}
\begin{lstlisting}[style=prompt]
Wrong code:
```
data = nl.zeros((par_dim(128), 128), dtype=np.float32)

for i in nl.sequential_range(4):
  i_tile = nisa.iota(i, dtype=nl.uint32).broadcast_to(data.shape)
  data = data + i_tile # Warning: shadowing local tensor 'float32 data[128, 128]' with a new object, use 'data[...] =' if you want to update the existing object

nl.store(ptr, value=data) # # Error: Local variable 'tmp' is referenced outside of its parent scope ...
```

Correct code:
```
data = nl.zeros((par_dim(128), 128), dtype=np.float32)

for i in nl.sequential_range(4):
  i_tile = nisa.iota(i, dtype=nl.uint32).broadcast_to(data.shape)
  data[...] = data + i_tile

nl.store(ptr, value=data)
```

## Access variables
1. Don't use slice with variable size
2. List indices must be integers or slices, not Index
3. Shape element must be integers
4. InstTile cannot be directly assigned to a tensor, use store operation instead.


# Problem
```
{problem_code}
```

# Baseline NKI kernel
```
{kernel_code}
```

# Kernel usage
```
if __name__ == "__main__":
    inputs = get_inputs()
    ref_output = forward(*inputs)
    kernel_output = transform_nki_outputs(kernel(*transform_to_nki_inputs(inputs)), ref_output)
    assert np.allclose(kernel_output, ref_output, atol=1e-4, rtol=1e-2)
```

# Profile
```
{profile}
```

Output the optimized `kernel` function wrapped in code block.

\end{lstlisting}
\end{minipage}
\caption{Base prompt for sampling Claude Sonnet 4. The problem\_code, kernel\_code, and profile will be replaced with the actual values.}
\label{lst:claude-prompt-c}
\end{figure*}
%%%%%%%%%%%%%%%%%%%%%%%%%%%%%%%%%%%%%%%%%%%%%%%%%%%%%%%%%%%%%%%%%%%%%%%%%%%%%%%
%%%%%%%%%%%%%%%%%%%%%%%%%%%%%%%%%%%%%%%%%%%%%%%%%%%%%%%%%%%%%%%%%%%%%%%%%%%%%%%

\end{document}